# PCANet: An energy perspective

Jiasong Wu, *Member, IEEE*, Shijie Qiu, Youyong Kong, Longyu Jiang, Lotfi Senhadji, *Senior Member, IEEE*, Huazhong Shu, *Senior Member, IEEE*

*Abstract*—The principal component analysis network (PCANet), which is one of the recently proposed deep learning architectures, achieves the state-of-the-art classification accuracy in various databases. However, the explanation of the PCANet is lacked. In this paper, we try to explain why PCANet works well from energy perspective point of view based on a set of experiments. The impact of various parameters on the error rate of PCANet is analyzed in depth. It was found that this error rate is correlated with the logarithm of energy of image. The proposed energy explanation approach can be used as a testing method for checking if every step of the constructed networks is necessary.

## I. INTRODUCTION

Deep learning [1-7], especially convolutional neural networks (CNNs) [8, 9], is a hot spot topic and it achieves the state-of-the-art results in many image database classification, including ImageNet large scale visual recognition [10-14], LFW face recognition [15-17], and handwritten digit recognition [8, 18], etc. The great success of deep learning systems is impressive, but a fundamental question still remains: Why do they work [19]? In the recent years, several attempts have been made for explaining the deep learning systems. These attempts can be roughly categorized into six groups: *1) Renormalization Theory*. Mehta and Schwab [20] constructed an exact mapping from the variational renormalization group (RG) scheme [21] to deep neural networks (DNNs) based on Restricted Boltzmann Machines (RBMs) [1, 2], and thus explained DNNs as a RG-like procedure to extract relevant features from structured data. *2) Probabilistic Theory*. Patel *et al.* [19] developed a new probabilistic framework for deep learning based on a Bayesian generative probabilistic model. By relaxing the generative model to a discriminative one, their models recover two of the current leading deep learning systems: deep CNNs and random decision forests (RDFs). *3) Information Theory.* Tishby and Zaslavsky [22] analyzed DNNs via the theoretical framework of the information bottleneck principle. Steeg and Galstyan [23] further introduced a framework for unsupervised learning of deep representations based on a hierarchical decomposition of information. *4)Developmental robotic perspective.* Sigaud and Droniou [24] scrutinized deep learning techniques under the light of their capability to construct a hierarchy of multimodal representations from the raw sensors of robots. *5)Group Theory*. Paul and Venkatasubramanian [25] explained deep learning system from the group-theoretic perspective point of view and showed why higher layers of deep learning framework tend to learn more abstract features. Shaham *et al.* [26] discussed the approximation of wavelet functions using deep neural nets. Anselmi *et al*. [27] explained deep CNNs by invariant and selective theory, whose ideas come from *i-*Theory [28], which is a recent theory of feedforward processing in sensory cortex. *6) Energy perspective.* The mathematical analysis of CNNs was performed by Mallat [29], in which wavelet scattering network (ScatNet) was proposed. The convolutional layer, nonlinear layer, pooling layer were constructed by prefixed complex wavelets, modulus operator, and average operator, respectively. **Owning to its characteristic of using prefixed filters which are not learned from data,** ScatNet was explained in [29] from *energy perspective* both in theory and experiment aspect, that is, **ScatNet keeps the energy of image in each layer although using modulus operator.** ScatNet achieves the state-of-the-art results in various image classification tasks [30] and was then extended to semi-discrete frame networks [31] as well as complex valued convolutional nets [32].

On the other hand, Chan *et al*. [33] recently proposed a new deep learning algorithm called principal component analysis network (PCANet), whose convolutional layer, nonlinear processing layer, and feature pooling layer consist of principal component analysis (PCA) filter bank, binarization, and block-wise histogram, respectively. Although PCANet is constructed with most basic units, it surprisingly achieves the state-of-the-art performance for most image classification tasks. PCANet arouses the interest of many researchers in this field. For example, Gan *et al*. proposed a compressive sensing network (CSNet) [34] and a graph embedding network (GENet) [35] for image classification. Feng *et al*. [36] presented a discriminative locality alignment network (DLANet) for scene classification. Ng and Teoh [37] proposed discrete cosine transform network (DCTNet) for face recognition. Gan *et al*. [38] presented a PCA-based convolutional network by

This work was supported by the National Natural Science Foundation of China under Grants 61201344, 61271312,61401085, 81101104,11301074 and 61073138, by the Research Fund for the Doctoral Program of Higher Education (No. 20120092120036), the Project-sponsored by SRF for ROCS, SEM, and by Natural Science Foundation of Jiangsu Province under Grants BK20150650, DZXX-031, BY2014127-11, by the '333' project under Grant BRA2015288, by the Qing Lan Project, and by the High-end Foreign Experts Recruitment Program (GDT20153200043).

Jiasong Wu, Shijie Qiu, Youyong Kong, Longyu Jiang, and Huazhong Shu are with the LIST, Key Laboratory of Computer Network and Information Integration (Southeast University), Ministry of Education, and also with Centre de Recherche en Information Biomédicale sino-français, 210096 Nanjing, China (e-mail: jswu@seu.edu.cn, 1063392760@qq.com, kongyouyong@seu.edu.cn, jly@seu.edu.cn, shu.list@seu.edu.cn).

Lotfi Senhadji is with the INSERM, U 1099, Rennes 35000, France, and Université de Rennes 1, LTSI, Rennes 35000, France, and with Centre de Recherche en Information Biomédicale Sino-français, Rennes 35000, France (e-mail: lotfi.senhadji@univ-rennes1.fr)

combining the structure of PCANet and the LeNet-5 [8, 9]. Zeng *et al*. [39] proposed a multilinear discriminant analysis network (MLDANet) for tensor object classification. Although PCANet has been extensively investigated, the question still remains: Why it works well by using the most basic and simple units? To the best of our knowledge, the attempt to explain the PCANet [33] has not been reported in the literature.

Inspired by the research work proposed by Mallat and Bruna [29, 30], in this paper, we try to explain every step of PCANet in terms of energy perspective. The most difference between PCANet and ScatNet [29, 30] is that the convolutional layer of PCANet is constructed by data-dependent PCA filter while the convolutional layer of ScatNet is constructed by data-free complex wavelet filter. In this paper, we explain PCANet from an *energy perspective* on experiment aspect by using four image databases: Yale database [40], AR database [41], CMU PIE face database [42], and ORL database [43].

The paper is organized as follows. PCANet is reviewed in section II. Section III analyzes the impact of various parameters on the error rate of PCANet in depth and also gives the energy explanation of PCANet. Section IV concludes the work.

## II. REVIEW OF PRINCIPAL COMPONENT ANALYSIS NETWORK

In this section, we first review the PCANet [33], whose architecture is shown in Fig. 1 and can be divided into three stages, including 10 steps. Suppose that we have $N$ input training images $\{\mathbf{I}_i\}_{i=1}^{N}$ of size $m \times n$. Assuming that the patch size (or two-dimensional filter size) of all stages is $k_1 \times k_2$, where $k_1$ and $k_2$ are odd integers and satisfy $1 \leq k_1 \leq m$, $1 \leq k_2 \leq n$. We further assume that the number of filters in layer $i$ is $L_i$, that is, $L_1$ for the first stage and $L_2$ for the second stage, and so on. In the following, we describe the structure of PCANet in detail.

Let the $N$ input images $\{\mathbf{I}_i \in \mathbb{R}^{m \times n}\}$ be concatenated as follows:
$$\mathbf{I} = [\mathbf{I}_1 \quad \mathbf{I}_2 \quad \cdots \quad \mathbf{I}_N] \in \mathbb{R}^{m \times Nn}. \tag{1}$$

### A. The first stage of PCANet

As shown in Fig. 1, the first stage of PCANet includes the following 3 steps:

*Step 1: the first patch sliding process.*

We use a patch of size $k_1 \times k_2$ to slide each pixel of the $i$th image $\mathbf{I}_i \in \mathbb{R}^{m \times n}$ and then reshape each $k_1 \times k_2$ matrix into a column vector, which is then concatenated to obtain a matrix
$$\mathbf{X}_i = [\mathbf{x}_{i,1} \quad \mathbf{x}_{i,2} \quad \cdots \quad \mathbf{x}_{i,mn}] \in \mathbb{R}^{k_1 k_2 \times mn}, i = 1, 2, ..., N, \tag{2}$$
where $\mathbf{x}_{i,j}$ denotes the $j$th vectorized patch in $\mathbf{I}_i$.

Therefore, for all the input training images $\{\mathbf{I}_i\}_{i=1}^{N}$, we can obtain the following matrix
$$\mathbf{X} = [\mathbf{X}_1 \quad \mathbf{X}_2 \quad \cdots \quad \mathbf{X}_N] \in \mathbb{R}^{k_1 k_2 \times Nmn}, \tag{3}$$

*Step 2: the first mean remove process.*

In this step, we subtract patch mean from each patch and obtain
$$\overline{\mathbf{X}}_i = [\overline{\mathbf{x}}_{i,1} \quad \overline{\mathbf{x}}_{i,2} \quad \cdots \quad \overline{\mathbf{x}}_{i,mn}] \in \mathbb{R}^{k_1 k_2 \times mn}, i = 1, 2, ..., N, \tag{4}$$

where $\overline{\mathbf{x}}_{i,j} = \mathbf{x}_{i,j} - \frac{1}{mn}\sum_{j=1}^{mn} \mathbf{x}_{i,j}$, is a mean-removed vector.

For each input training image $\mathbf{I}_i \in \mathbb{R}^{m \times n}$, we can get a substituted matrix $\overline{\mathbf{X}}_i \in \mathbb{R}^{k_1 k_2 \times mn}$. Thus, for all the input training images $\{\mathbf{I}_i\}_{i=1}^{N}$, we can obtain the following matrix
$$\overline{\mathbf{X}} = [\overline{\mathbf{X}}_1 \quad \overline{\mathbf{X}}_2 \quad \cdots \quad \overline{\mathbf{X}}_N] \in \mathbb{R}^{k_1 k_2 \times Nmn}. \tag{5}$$

*Step 3: the first PCA process.*

In this step, we get the eigenvalues and eigenvectors of $\overline{\mathbf{X}}$ in (5) by using PCA algorithm, which in fact minimizes the reconstruction error in Frobenius norm as follows:
$$\min_{\mathbf{U} \in \mathbb{R}^{k_1 k_2 \times L_1}} \|\overline{\mathbf{X}} - \mathbf{U}\mathbf{U}^T \overline{\mathbf{X}}\|_F^2, \quad \text{s.t.} \quad \mathbf{U}^T \mathbf{U} = \mathbf{I}_{L_1}, \tag{6}$$
where $\mathbf{I}_{L_1}$ is an identity matrix of size $L_1 \times L_1$. Eq. (6) can be solved by eigenvalue decomposition method:
$$\mathbf{C}_1 = \mathbf{U}\mathbf{\Lambda}_1 \mathbf{U}^T, \tag{7}$$
where $\mathbf{C}_1$ is a covariance matrix given by
$$\mathbf{C}_1 = \frac{1}{Nmn} \overline{\mathbf{X}}\overline{\mathbf{X}}^T \in \mathbb{R}^{k_1 k_2 \times k_1 k_2}, \tag{8}$$
and $\mathbf{\Lambda}_1$ is a diagonal matrix composed of the first $L_1$ largest eigenvalues of $\mathbf{C}_1$,
$$\mathbf{\Lambda}_1 = \begin{bmatrix} \lambda_1^1 & & & \\ & \lambda_2^1 & & \\ & & \ddots & \\ & & & \lambda_{L_1}^1 \end{bmatrix}, \tag{9}$$
where $\lambda_1^1 \geq \lambda_2^1 \geq \cdots \geq \lambda_{L_1}^1$, and the italic superscript *1* denotes the first stage, and $\mathbf{U}$ is the matrix composed of the first $L_1$ principal eigenvectors
$$\mathbf{U} = [\mathbf{u}_1 \quad \mathbf{u}_2 \quad \cdots \quad \mathbf{u}_{L_1}] \in \mathbb{R}^{k_1 k_2 \times L_1}. \tag{10}$$

The PCA filter of the first stage of PCANet is then obtained from (10) by
$$\mathbf{W}_l^1 = \text{mat}_{k_1, k_2}(\mathbf{u}_l) \in \mathbb{R}^{k_1 \times k_2}, l = 1, 2, ..., L_1, \tag{11}$$
where $\text{mat}_{k_1, k_2}(\mathbf{u}_l)$ is a function that maps $\mathbf{u}_l \in \mathbb{R}^{k_1 k_2}$ to a matrix $\mathbf{W}_l^1 \in \mathbb{R}^{k_1 \times k_2}$.

The output of the first stage of PCANet is given by
$$\mathbf{I}_{i,l}^1 = \mathbf{I}_i * \mathbf{W}_l^1 \in \mathbb{R}^{m \times n}, \quad l = 1, 2, ..., L_1; \quad i = 1, 2, ..., N, \tag{12}$$
where $*$ denotes two-dimensional (2-D) convolution, and the boundary of $\mathbf{I}_i$ is zero-padded before convolving with $\mathbf{W}_l^1$ in order to make $\mathbf{I}_{i,l}^1$ having the same size as $\mathbf{I}_i$.

For each input image $\mathbf{I}_i \in \mathbb{R}^{m \times n}$, we obtain $L_1$ output images $\{\mathbf{I}_{i,l}^1, l = 1, 2, ..., L_1\} \in \mathbb{R}^{m \times n}$ after the first stage of PCANet. We denote $\mathbf{I}^1$ as
$$\mathbf{I}^1 = [\mathbf{I}_{1,1}^1 \quad \cdots \quad \mathbf{I}_{1,L_1}^1 \quad \cdots \quad \mathbf{I}_{N,1}^1 \quad \cdots \quad \mathbf{I}_{N,L_1}^1] \in \mathbb{R}^{m \times NL_1 n}. \tag{13}$$

### B. The second stage of PCANet

As shown in Fig. 1, the second stage of PCANet also includes 3 steps:

*Step 4: the second patch sliding process.*

Similar to *Step 1*, we use a patch of size $k_1 \times k_2$ to slide each pixel of the *i*th image $\mathbf{I}_{i,l}^1 \in \mathbb{R}^{k_1 \times k_2}, l=1,2,...,L_1$, and obtain a matrix

$$\mathbf{Y}_{i,l} = \begin{bmatrix} \mathbf{y}_{i,l,1} & \mathbf{y}_{i,l,2} & \cdots & \mathbf{y}_{i,l,mn} \end{bmatrix} \in \mathbb{R}^{k_1 k_2 \times mn}, \quad (14)$$
$$l=1,2,...,L_1; \; i=1,2,...,N,$$

where $\mathbf{y}_{i,j,l}$ denotes the *j*th vectorized patch in $\mathbf{I}_{i,l}^1$.

Therefore, for *i*th filter all the input training images $\{\mathbf{I}_{i,l}^1\}_{i=1}^{N}$, we can obtain the following matrix

$$\mathbf{Y}_l = \begin{bmatrix} \mathbf{Y}_{1,l} & \mathbf{Y}_{2,l} & \cdots & \mathbf{Y}_{N,l} \end{bmatrix} \in \mathbb{R}^{k_1 k_2 \times Nmn}, \quad (15)$$

We concatenate the matrices of all the $L_1$ filters and obtain

$$\mathbf{Y} = \begin{bmatrix} \mathbf{Y}_1 & \mathbf{Y}_2 & \cdots & \mathbf{Y}_{L_1} \end{bmatrix} \in \mathbb{R}^{k_1 k_2 \times L_1 Nmn}. \quad (16)$$

*Step 5: the second mean remove process.*

Similar to *Step 2*, we obtain the mean-removed version of (16) by

$$\overline{\mathbf{Y}} = \begin{bmatrix} \overline{\mathbf{Y}}_1 & \overline{\mathbf{Y}}_2 & \cdots & \overline{\mathbf{Y}}_{L_1} \end{bmatrix} \in \mathbb{R}^{k_1 k_2 \times L_1 Nmn}, l=1,2,...,L_1, \quad (17)$$

where

$$\overline{\mathbf{Y}}_l = \begin{bmatrix} \overline{\mathbf{Y}}_{1,l} & \overline{\mathbf{Y}}_{2,l} & \cdots & \overline{\mathbf{Y}}_{N,l} \end{bmatrix} \in \mathbb{R}^{k_1 k_2 \times Nmn}, \quad (18)$$

$$\overline{\mathbf{Y}}_{i,l} = \begin{bmatrix} \overline{\mathbf{y}}_{i,l,1} & \overline{\mathbf{y}}_{i,l,2} & \cdots & \overline{\mathbf{y}}_{i,l,mn} \end{bmatrix} \in \mathbb{R}^{k_1 k_2 \times mn}, \quad (19)$$
$$l=1,2,...,L_1; \; i=1,2,...,N,$$

$$\overline{\mathbf{y}}_{i,l,j} = \mathbf{y}_{i,l,j} - \frac{1}{mn}\sum_{j=1}^{mn}\mathbf{y}_{i,l,j}, l=1,2,...,L_1; \; i=1,2,...,N. \quad (20)$$

*Step 6: the second PCA process.*

Similar to *Step 3*, we use the PCA algorithm to minimize the following optimization problem:

$$\min_{\mathbf{V} \in \mathbb{R}^{k_1 k_2 \times L_2}} \|\overline{\mathbf{Y}} - \mathbf{V}\mathbf{V}^T\overline{\mathbf{Y}}\|_F^2, \text{ s.t. } \mathbf{V}^T\mathbf{V} = \mathbf{I}_{L_2}, \quad (21)$$

where $\mathbf{I}_{L_2}$ is an identity matrix of size $L_2 \times L_2$. Eq. (21) can be solved by eigenvalue decomposition method:

$$\mathbf{C}_2 = \mathbf{V}\mathbf{\Lambda}_2\mathbf{V}^T, \quad (22)$$

where

$$\mathbf{C}_2 = \frac{1}{L_1 Nmn}\overline{\mathbf{Y}}\overline{\mathbf{Y}}^T \in \mathbb{R}^{k_1 k_2 \times k_1 k_2}, \quad (23)$$

$$\mathbf{\Lambda}_2 = \begin{bmatrix} \lambda_1^2 & & & \\ & \lambda_2^2 & & \\ & & \ddots & \\ & & & \lambda_{L_2}^2 \end{bmatrix}, \lambda_1^2 \geq \lambda_2^2 \geq \cdots \geq \lambda_{L_2}^2. \quad (24)$$

$$\mathbf{V} = \begin{bmatrix} \mathbf{v}_1 & \mathbf{v}_2 & \cdots & \mathbf{v}_{L_2} \end{bmatrix} \in \mathbb{R}^{k_1 k_2 \times L_2}. \quad (25)$$

Note that the italic superscript *2* in (24) denotes the second stage. The PCA filter of the second stage of PCANet is then obtained from (25) by

$$\mathbf{W}_\ell^2 = \text{mat}_{k_1,k_2}(\mathbf{v}_\ell) \in \mathbb{R}^{k_1 \times k_2}, \ell=1,2,...,L_2, \quad (26)$$

where $\text{mat}_{k_1,k_2}(\mathbf{v}_\ell)$ is a function that maps $\mathbf{v}_\ell \in \mathbb{R}^{k_1 k_2}$ to a matrix $\mathbf{W}_\ell^2 \in \mathbb{R}^{k_1 \times k_2}$. Therefore, the output of the second stage of PCANet is given by

$$\mathbf{I}_{i,l,\ell}^2 = \mathbf{I}_{i,l}^1 * \mathbf{W}_\ell^2 \in \mathbb{R}^{k_1 \times k_2}, \; l=1,2,...,L_1; \quad (27)$$
$$\ell=1,2,...,L_2; \; i=1,2,...,N,$$

where $*$ denotes 2-D convolution.

For each input image $\mathbf{I}_{i,l}^1 \in \mathbb{R}^{m \times n}$, we obtain $L_2$ output images $\{\mathbf{I}_{i,l,\ell}^2, \ell=1,2,...,L_2\} \in \mathbb{R}^{m \times n}$ after the second stage of PCANet. Thus, we obtain $NL_1 L_2$ images $\{\mathbf{I}_{i,l,\ell}^2 \in \mathbb{R}^{m \times n}\}, l=1,2,...,L_1; \ell=1,2,...,L_2; i=1,2,...,N$, for all the input training images $\{\mathbf{I}_i \in \mathbb{R}^{m \times n}\}, i=1,2,...,N$ after the first and second stages.

We denote $\mathbf{I}^2$ as

$$\mathbf{I}^2 = \begin{bmatrix} \mathbf{I}_{1,1,1}^2 & \cdots & \mathbf{I}_{1,1,L_2}^2 & \cdots & \mathbf{I}_{1,L_1,1}^2 & \cdots & \mathbf{I}_{1,L_1,L_2}^2 & \cdots \\ \cdots & \mathbf{I}_{N,1,1}^2 & \cdots & \mathbf{I}_{N,1,L_2}^2 & \cdots & \mathbf{I}_{N,L_1,1}^2 & \cdots & \mathbf{I}_{N,L_1,L_2}^2 \end{bmatrix}. \quad (28)$$

*C. The output stage of PCANet*

*Step 7: binary quantization.*

In this step, we binarize the outputs $\mathbf{I}_{i,l,\ell}^2$ of the second stage of PCANet and obtain

$$\mathbf{P}_{i,l,\ell} = H(\mathbf{I}_{i,l,\ell}^2), l=1,2,...,L_1; \; \ell=1,2,...,L_2; \; i=1,2,...,N, \quad (29)$$

where $H(\cdot)$ is a Heaviside step function whose value is one for positive entries and zero otherwise. We denote $\mathbf{P}$ as

$$\mathbf{P} = \begin{bmatrix} \mathbf{P}_{1,1,1} & \cdots & \mathbf{P}_{1,1,L_2} & \cdots & \mathbf{P}_{1,L_1,1} & \cdots & \mathbf{P}_{1,L_1,L_2} & \cdots \\ \cdots & \mathbf{P}_{N,1,1} & \cdots & \mathbf{P}_{N,1,L_2} & \cdots & \mathbf{P}_{N,L_1,1} & \cdots & \mathbf{P}_{N,L_1,L_2} \end{bmatrix}. \quad (30)$$

*Step 8: weight and sum.*

Around each pixel, we view the vector of $L_2$ binary bits as a decimal number. This converts the binary images $\{\mathbf{P}_{i,l,\ell}\}$ back into integer-valued images as follows:

$$\mathbf{T}_{i,l} = \sum_{\ell=1}^{L_2} 2^{\ell-1} \mathbf{P}_{i,l,\ell}, \quad (31)$$

We denote $\mathbf{T}$ as

$$\mathbf{T} = \begin{bmatrix} \mathbf{T}_{1,1} & \cdots & \mathbf{T}_{1,L_1} & \cdots & \mathbf{T}_{N,1} & \cdots & \mathbf{T}_{N,L_1} \end{bmatrix} \in \mathbb{R}^{m \times NL_1 n}. \quad (32)$$

*Step 9: block sliding.*

We use a block of size $h_1 \times h_2$ to slide each of the $L_1$ images $\mathbf{T}_{i,l}, l=1,...,L_1$, with overlap ratio $R$, and then reshape each $h_1 \times h_2$ matrix into a columnvector, which is then concatenated to obtain a matrix

$$\mathbf{Z}_{i,l} = \begin{bmatrix} \mathbf{z}_{i,l,1} & \mathbf{z}_{i,l,2} & \cdots & \mathbf{z}_{i,l,B} \end{bmatrix} \in \mathbb{R}^{h_1 h_2 \times B}, i=1,2,...,N, \quad (33)$$

where $\mathbf{z}_{i,l,j}$ denotes the *j*th vectorized patch in $\mathbf{T}_{i,l}, l=1,...,L_1$. $B$ is the number of blocks when using a block of size $h_1 \times h_2$ to slide each $\mathbf{T}_{i,l}, l=1,...,L_1$, with overlap ratio $R$.

For $L_1$ images, we concatenate $\mathbf{Z}_{i,l}$ to obtain a matrix

$$\mathbf{Z}_i = \begin{bmatrix} \mathbf{Z}_{i,1} & \mathbf{Z}_{i,2} & \cdots & \mathbf{Z}_{i,B} \end{bmatrix} \in \mathbb{R}^{h_1 h_2 \times L_1 B}, i=1,2,...,N, \quad (34)$$

We denote $\mathbf{Z}$ as

$$\mathbf{Z} = \begin{bmatrix} \mathbf{Z}_{1,1} & \cdots & \mathbf{Z}_{1,B} & \cdots & \mathbf{Z}_{N,1} & \cdots & \mathbf{Z}_{N,B} \end{bmatrix} \in \mathbb{R}^{h_1 h_2 \times L_1 BN}. \quad (35)$$

*Step 10: histogram.*

We compute the histogram (with $2^{L_2}$ bins) of the decimal values in each column of $\mathbf{Z}_i$ and concatenate all the histograms into one vector and obtain

$$\mathbf{f}_i = \begin{bmatrix} \text{Hist}(\mathbf{z}_{i,1,1}) & \cdots & \text{Hist}(\mathbf{z}_{i,1,B}) & \cdots \\ \cdots & \text{Hist}(\mathbf{z}_{i,L_1,1}) & \cdots & \text{Hist}(\mathbf{z}_{i,L_1,B}) \end{bmatrix}^T \in \mathbb{R}^{(2^{L_2})L_1B}. \quad (36)$$

which is the "feature" of the input image $\mathbf{I}_i$ and Hist(.) denotes the histogram operation. We denote $\mathbf{f}$ as

$$\mathbf{f} = \begin{bmatrix} \mathbf{f}_1 & \cdots & \mathbf{f}_N \end{bmatrix} \in \mathbb{R}^{(2^{L_2})L_1BN}. \quad (37)$$

*D. The parameters of PCANet*

Table I lists the parameters of PCANet and their descriptions. For a given dataset, $m$ and $n$ are fixed. In this paper, we always set $k_1=k_2=3$ due to the following reasons: (1) we have $1 \leq L_1 \leq 9$ and $1 \leq L_2 \leq 9$, that is to say, the number of outputs of PCANet is at most 81 times of that of input after the *Step 6*. The consumption of memory is tolerable for an ordinary computer (for example, 16 GB RAM). (2) We can exhaust the combination of $L_1$ and $L_2$ in our experiments since the total number of combinations of $L_1$ and $L_2$ is 81. After the above setting, there are five free parameters left: $L_1$, $L_2$, $h_1$, $h_2$, and $R$.

## III. ERROR RATE ANALYSIS AND ENERGY EXPLANATION OF PCANET

In this section, we perform error rate analysis for PCANet. The following experiments were implemented in Matlab programming language on a PC machine, which sets up Microsoft Windows 7 operating system and has an Intel(R) Core(TM) i7-2600 CPU with speed of 3.40GHz and 16GB RAM.

*A. Experiment databases*

In the following, we use four databases, whose matlab format is provided in [44], to analyze the error rate change as well as the energy change of PCANet: **1)Yale database** [40]. It contains 165 grayscale images (15 individuals, 11 images per individual), which are randomly divided into 30 training images and 135 testing images; **2)AR database** [41]. We use a non-occluded subset of AR database containing 700 images (50 male subjects, 14 images per subject), which are randomly divided into 200 training images and 500 testing images; **3)The CMU PIE face database** [42]. We use pose C27 (a frontal pose) subset of CMU PIE face database which has 3400 images (68 persons, 50 images per person), which are randomly divided into 1000 training images and 2400 testing images; **4)ORL database** [43]. It contains 400 grayscale images (40 individuals, 10 images per individual), which are randomly divided into 80 training images and 320 testing images. Note that the size of all the images is 32×32 pixels. Figs. S1, S4, S10, and S14 show the image example of four databases. **Fig. S1 means the first figure of supplement document**.

*B. Experiment methods*

In this section, we use the above mentioned databases to analyze the PCANet by using energy method. We first introduce the definition of the energy of a two-dimensional image $\mathbf{I}(i,j)$ of size $m \times n$:

$$E(\mathbf{I}) = \sum_{i=1}^{m} \sum_{j=1}^{n} \mathbf{I}^2(i,j). \quad (38)$$

Fig. 1 shows that we record 10 energies for PCANet. It is convenient for us to divide these energies into two parts: Energies 1-9 belong to the first part and Energy 10 belongs to the second part. The reason for this classification is that Energies 1-9 are not related to the block size $h_1 \times h_2$ and the overlap ratio $R$. The values that we recorded in the whole process of PCANet are shown in Table II, which includes 10 energies and the error rate. Note that we do not take the energy of the feature $\mathbf{f}$ in (37) into consideration, since the block-wise histogram is only the statistic (or occurrent frequency) of energy value and we consider that this process does not change the energy.

**(1) The second part**

In this subsection, we first consider the second part since it is much closer to error rate $e$ than the first part. In the second part, there are five parameters: $h_1$ (the number of rows of block), $h_2$ (the number of columns of block), $R$ (the overlap ratio), $L_1$ (the number of filters of the 1st stage), $L_2$ (the number of filters of the 2nd stage). Meanwhile, we can also obtain the BlockEnergy (Energy 10) corresponding to each error rate $e$ for each combination of $h_1$, $h_2$, $R$, $L_1$, and $L_2$. Therefore, the aim of this subsection is to explore qualitatively the relationships between the error rate $e$, the BlockEnergy and the parameters $h_1$, $h_2$, $R$, $L_1$, $L_2$.

1) How the error rate $e$ varies according to $h_1$, $h_2$, $R$, $L_1$, and $L_2$?

In this experiment, we use Yale database for analyzing the error rate $e$ in relation with $h_1$, $h_2$, $R$, $L_1$, $L_2$, where $h_1$, $h_2 \in \{1:1:32\}$, $L_1$, $L_2 \in \{1:1:9\}$, and $R \in \{0:0.1:0.9\}$. Note that $\{a:b:c\}$ means the set contains the values that vary from $a$ to $c$ with increment $b$. The results are shown in Fig. 2, whose vertical axis is $h_1 \in \{1:1:32\}$, and horizontal axis is related to $h_2$, $R$, and $L_2$. Specifically, the horizontal axis can be divided into 9 blocks corresponding to $L_2 \in \{1:1:9\}$, each block can be divided into 10 sub-blocks corresponding to $R \in \{0:0.1:0.9\}$, and each sub-block has the size of 32×32 corresponding to $h_2 \in \{1:1:32\}$. Therefore, the range of horizontal axis value is from 1 to 2880 (=32×10×9). Fig. 2(*a*) is in fact the concatenated version of 90 images, which are provided in Fig. S2 of the supplement document. Fig. 3 shows 18 images of them. In Figs. 2 and 3, the error rate $e$ is expressed by colors. The brown and the blue denote the error rate 1 and 0, respectively. The pure brown area appears in the figure in fact denotes that the algorithm cannot work in these cases, specifically, **the function of "HashingHist" in the PCANet software can not deal with these cases, we manually set the error rate to 1**. It does not matter, these are just used in Fig. 2 as separators of the results of various $L_2$. From Figs. 2 and 3, we can see that:

a)*The impact of $L_1$ on the error rate $e$*. Fig. 2 contains 9 similar figures corresponding to $L_1 \in \{1:1:9\}$. If we fix $L_2$, we can see that the error rate $e$ is gradually decreased when $L_1$ increasing. When $L_1 \geq 4$, the figures are gradually stable. The first three figures have notable changes, this phenomena implies that the first few number of filters captures most of the energies of the original image, leading to the severe changes of error rates.

b)*The impact of $L_2$ on the error rate $e$*. Let us first fix $R=0$ and see the 9 figures ($a_{L_2,0}$, $L_2 \in \{1:1:9\}$) in Fig. 3, whose the origin is located at the left top point, and the horizontal axis and the vertical axis are $h_2$ and $h_1$, respectively. Every figure can be divided into four blocks: left up block, right up block, left down block, and right down block. We can see that the separate lines for the four blocks are the same for 9 figures, however, the blue area (corresponds to the low error rate area) is gradually deeper in terms of the increased direction of $L_2$. Moreover, as the value of $L_2$ increases, the first six figures corresponding to $L_2 \in \{1:1:6\}$ change a lot, but when $L_2 \geq 7$, the left three figures only make little changes. Therefore, we can think that when $L_2 = 7$, the error rate is stable. For other choice of $R \in \{0.1:0.1:0.9\}$, the similar phenomenon can be observed.

Therefore, we can conclude that **$L_2$ influences directly the area colors (corresponds to error rate) of four blocks. In general, the bigger $L_2$, the more blue area (or the lower error rate). However, the change is minor when $L_2 \geq 7$.**

c) *The impact of $R$ on the error rate $e$*. Let us first fix $L_2=1$ and see the 10 figures ($a_{0,R}$, $R \in \{0:0.1:0.9\}$) in Fig. 3. We can see that the error rate in the left up block is better than that of other three blocks in general. For a fixed $L_2$, the blue area (the low error rate area) is expanding from northwest corner to southeast corner in terms of the increased direction of overlap ratio $R$. The separator lines are different for different $R$, however, the most blue area (or the lower error rate) is always appeared in the left up block. We can also see that the appropriate values of $R$ belong to $\{0:0.1:0.6\}$. For other choice of $L_2 \in \{2:1:9\}$, the similar phenomenon can be seen.

Therefore, we can conclude that **$R$ influences directly the area size of four blocks in terms of different separator lines. Since the error rate of the left up block is lower than other three blocks, therefore, $R$ indirectly influences the error rate of PCANet.**

d) *The impact of the block sizes $h_1$ and $h_2$ on the error rate $e$*. Can we easily find the block sizes $h_1$ and $h_2$ that achieve the least error rate $e_l$ in the block sliding process (Step 9) of PCANet? We do a further experiment to analyze the **least** error rate $e_l$ with $h_1$, $h_2$ and $R$. We fix $L_1=1$ and varies $L_2$ from 1 to 9, and varies $h_2$ from 1 to 32, and plot the value of $h_1$ that corresponds to the least error rate $e_l$ of all the overlap ratio $R$. The results, which is shown in Fig. S3 provided in the supplement document, show that the range of $h_1$ is very wide and the appearance of the least error rate $e_l$ seems not having a regular rule, that is, it is very difficult for us to search the **least** error rate $e_l$. Therefore, in the following of this paper, we set a constraint to the values of $h_1$ and $h_2$ and let $h_2 = \lfloor nh_1/m \rfloor$, that is to say, only choose the diagonal for searching a suboptimal error rate. The relative error of the least error rate of the diagonal $e_{la}$ and the least error rate of the whole image $e_l$ of Yale database are computed (The result is provided in Table S1 of the supplement document). The average relative error and the least relative error are 1.24 % and 0.74%, respectively. Therefore, in this paper, we set a constraint of $h_2 = \lfloor nh_1/m \rfloor$ is reasonable. In this case, the free parameters of PCANet are reduced from 5 to 4. Fig. 4 shows the error rate $e$ in relation with $R$, $L_1$, $L_2$ and $h_1$ ($h_2 = \lfloor nh_1/m \rfloor$ is implicit).

2) How the BlockEnergy (Energy 10) varies according to error rate $e$?

When we perform the PCANet on Yale dataset, we save the energy in every steps of Fig. 1. Fig. 5 shows the logarithm of BlockEnergy in relation with $R$, $L_1$, $L_2$ and $h_1$ ($h_2 = \lfloor nh_1/m \rfloor$ is implicit). Comparing Figs. 4 and 5, we can see that these two figures are very similar, that is to say, the appearance of the lower error rate $e$ is not random when changing the parameters, it is correlated with the logarithm of BlockEnergy (log(E(**Z**))).

In the following, we use the Curve Fitting Tool of Matlab to further investigate the correlation degree of the lower error rate $e$ with the logarithm of BlockEnergy(log(E(**Z**))). We fit the function $e = f(g)$, $g = 1/\log(E(\mathbf{Z}))$, by using the linear model Poly3 in Matlab with 95% confidence bounds, that is, we use $e = p_1 g^3 + p_2 g^2 + p_3 g + p_4$ to find the relationship of $e$ and $1/\log(E(\mathbf{Z}))$. The results of four databases are shown in Table III. The criterions include:

a) The error sum of squares (SSE), regression sum of squares (SSR), and total sum of squares (SST) are defined respectively as

$$SSE = \sum_{i=1}^{N}(e_i - f(g_i))^2 , \quad (39)$$

$$SSR = \sum_{i=1}^{N}(f(g_i) - \overline{e})^2 , \quad (40)$$

$$SST = SSE + SSR = \sum_{i=1}^{N}(e_i - \overline{e})^2 , \quad (41)$$

where $\overline{e} = \frac{1}{N}\sum_{i=1}^{N} e_i$, and $N=\max(L_1 L_2 h_1 R)= 9\times 9\times 32\times 10=25920$ in our experiment.

b) Coefficient of determination or $R$-square is defined as

$$R^2 = 1 - \frac{SSE}{SST} , \quad (42)$$

which is a number that indicates how well data fit a statistical model, for example, a curve.

c) Root mean squared error (RMSE) is defined as

$$RMSE = \sqrt{\frac{\sum_{i=1}^{N}(e_i - f(g_i))^2}{N}} . \quad (43)$$

As we can see from Table III, the $R$-square of the data of Yale database is 0.7696, which means 76.96% of the variance in the response variable can be explained by the explanatory variables. The remaining 23.04% can be attributed to unknown, lurking variables or inherent variability. Therefore, $e$ and log(E(**Z**)) are fairly strongly correlated. Similar conclusion can be obtained for other three databases.

*How to explain the block sliding process*? We can in fact explain this process as an "overlap filtering". The overlap ratio $R=0,0.1,0.2,...,0.9$, corresponds to the 1st, 2nd, 3rd, ..., 10th filter bank, respectively. Suppose we use a block $h_1 \times \lfloor nh_1/m \rfloor$, $1 \leq h_1 \leq m$, to slide an image, we can get at most 10 images. We compute the energy of the 10 images, and then rank the energies from the greatest to the smallest, which is shown in the 3th row of Table IV. We then show the ratio of energy in the 5th row. As shown in **Section b)**, $R \in \{0:0.1:0.6\}$ is appropriate for PCANet. We can see from Table IV, the energy we keep is about 95.75% when *R*=0.6, which seems the point that comprises the training error and the generalized error.

**(2) The first part**

In this subsection, we will analyze the changes of Energies 1-9 of the first part of PCANet shown in Fig. 1. Since the first part is not related to $h_1$, $h_2$, and *R*, so in the experiment of this section, we simply set $h_1 = h_2 = 8$, and *R*= 0.5. Let us take Yale dataset for example, Fig. 6 shows the energy values of every stages of the first part of PCANet versus the *n*th energy. The results of 9 filters, corresponding to $L_1 \in \{1:1:9\}$, are shown in different colors. For example, the lowest line is the result that we choose only the 1 largest eigenvector corresponding to $L_1$=1. The second lowest line is the result that we choose the 2 largest eigenvectors corresponding to $L_1$=2, and so on. Note that the vertical axis is shown in logarithmic coordinate. From Fig. 6, for all the 9 filters, we can easily see the energy changes are as follows

$$\begin{aligned}
\textbf{TrainEnergy} &\xrightarrow[\text{Step 1}]{+} \textbf{PatchEnergy}1 \\
&\xrightarrow[\text{Step 2}]{-} \textbf{PatchEnergyRed}1 \\
&\xrightarrow[\text{Step 3}]{+ \text{ or } -} \textbf{PCAEnergy}1 \xrightarrow[\text{Step 4}]{+} \textbf{PatchEnergy}2 \\
&\xrightarrow[\text{Step 5}]{\textcolor{red}{0}} \textbf{PatchEnergyRed}2 \\
&\xrightarrow[\text{Step 6}]{+} \textbf{PCAEnergy}2 \xrightarrow[\text{Step 7}]{-} \textbf{BinaryEnergy} \\
&\xrightarrow[\text{Step 8}]{+} \textbf{WeightSumEnergy},
\end{aligned} \quad (44)$$

where the arrow means that the energy values change from the left-hand side to the right-hand side. "+", "−", "0" above the arrow denote the energy increase, decrease, and equal process, respectively. Steps 1-8 below the arrow correspond to the eight steps of Fig. 1.

From (44), we can see that

1) For Steps 1 and 4, both of them are energy increase process, which can be easily understood since these two steps perform the overlapping patch sliding process. However, how can we explain the patch sliding process? In fact, the patch sliding process is a special case of block sliding process with the overlap ratio *R*=0.9 shown in **the second part**.

2) For the Steps 2 and 5, both of them are mean remove process. Step 2 is energy decrease process, which is reasonable since the mean remove process decreases the variation of images leading to the reduction of image energy. **However, it is strange that Step 5 keeps the energy. Therefore, we think this step may not be needed in the construction of PCANet.** We then perform some experiments on four databases to verify our inference. We set $L_1$=$L_2$=6, *R*=0.5, and $1 \leq h_1 = h_2 \leq 32$. Then, we compute the differences of error rates of PCANet with mean remove Step 5, denoted as $e_r$, and that of without Step 5, denoted as $e_{wr}$. The mean of differences of error rates for Yale database, AR database, CMU PIE database, ORL database are $4.6296 \times 10^{-4}$, $-1.8750 \times 10^{-4}$, $9.1146 \times 10^{-4}$, $-19.0 \times 10^{-4}$, respectively, all of which are very small.(The detailed values are provided in Table S3 of the supplement document). **Therefore, when taking the computational complexity into consideration, we recommend removing the mean remove process of Step 5.**

3) For the Steps 3 and 6, both of them are PCA process. Step 3 is energy increasing process except the case of 1 filter, but the energy change is very small. However, step 6 is energy decreasing process. From Fig. 6, we can see that the energy is even less than the original training energy after Step 6. Our explanation is that although two sliding process increases the energy through overlapping, however, the "added energy" is reduced to the level of training energy by the second PCA filter. Therefore, the second PCA filter is very important to the construction of PCANet since it reduces the added "pseudo energy". Table V shows the energy of every 9 filters (the 1st PCA), $h_1$=$h_2$=8，*R*=0.5，$L_2$=1, and $L_1$ varies from 1 to 9, and the energy of every 9 filters (the 2nd PCA), $h_1$=$h_2$=8，*R*=0.5，$L_1$=8, and $L_2$ varies from 1 to 9. Table VI shows the error rate corresponds to $L_1, L_2 \in \{1:1:9\}$ when $h_1$=$h_2$=8, *R*=0.5.

4) For the Steps 7 and 8, which cause the most sharp energy changes of PCANet. It is also reasonable since the binarization operation is a very powerful quantization process (nonlinear processing), which loses much energy. Meanwhile, the weighted and summed step is a very powerful energy increasing step. It is interesting to see from Fig. 6 that the energy in Step 8 gradually increases when $L_2$ varies from 1 to 9. Let's us focus on the two horizontaldashed lines: the first one denotes the **PatchEnergy2** which is the maximum energy of linear layer of PCANet, the second one denotes the **TrainEnergy**. We can see from Fig. 6 and Table VI:

a) When **WeightSumEnergy**<**TrainEnergy**, corresponding to Fig. 6 (*a*)-(*e*), the error rate is gradually decreased when $L_1 \in \{1:1:9\}$ and $L_2 \in \{1:1:5\}$.

b) When **TrainEnergy**<**WeightSumEnergy**≤ **PatchEnergy2**, corresponding to Fig. 6 (*f*)-(*h*), the error rate is gradually stable when $L_1 \in \{1:1:9\}$ and $L_2 \in \{6:1:8\}$.

c) When**WeightSumEnergy**> **PatchEnergy2**, corresponding to Fig. 6 (*i*), the error rate is increased when $L_1 \in \{1:1:9\}$ and $L_2$=9. In this case, we think the **WeightSumEnergy** includes "pseudo energy" so that the **WeightSumEnergy** is even larger than the maximum energy of linear layer of PCANet, that is, **PatchEnergy2.**

Note that the above experiments performed in other three databases (AR database, CMU PIE face database, ORL database), the results of which correspond to the Figs. 4-6, and Tables IV-VI of YALE database are provided in the supplement document due to the limited space. The results of other three databases are similar to that of Yale database.

From the energy change of PCANet, we can see that energy conservation of images by PCANet helps for obtaining good performances as for ScatNet [29, 30] method.

## IV. CONCLUSIONS

In this paper, the impact of various parameters on the error rate of PCANet has been analyzed in depth. It was found that this error rate is correlated with the logarithm of energy of images. The role of every step of PCANet has been investigated, for example, the patch sliding process and the block sliding process were explained by "overlap filtering". The proposed energy explanation approach could be used as a testing method for checking if every step of the constructed networks is necessary. For example, we find the second mean remove step is not needed in the construction of PCANet since the energy is not changed. Furthermore, the proposed energy explanation approach may be extended to other PCANet-based networks [34-39].

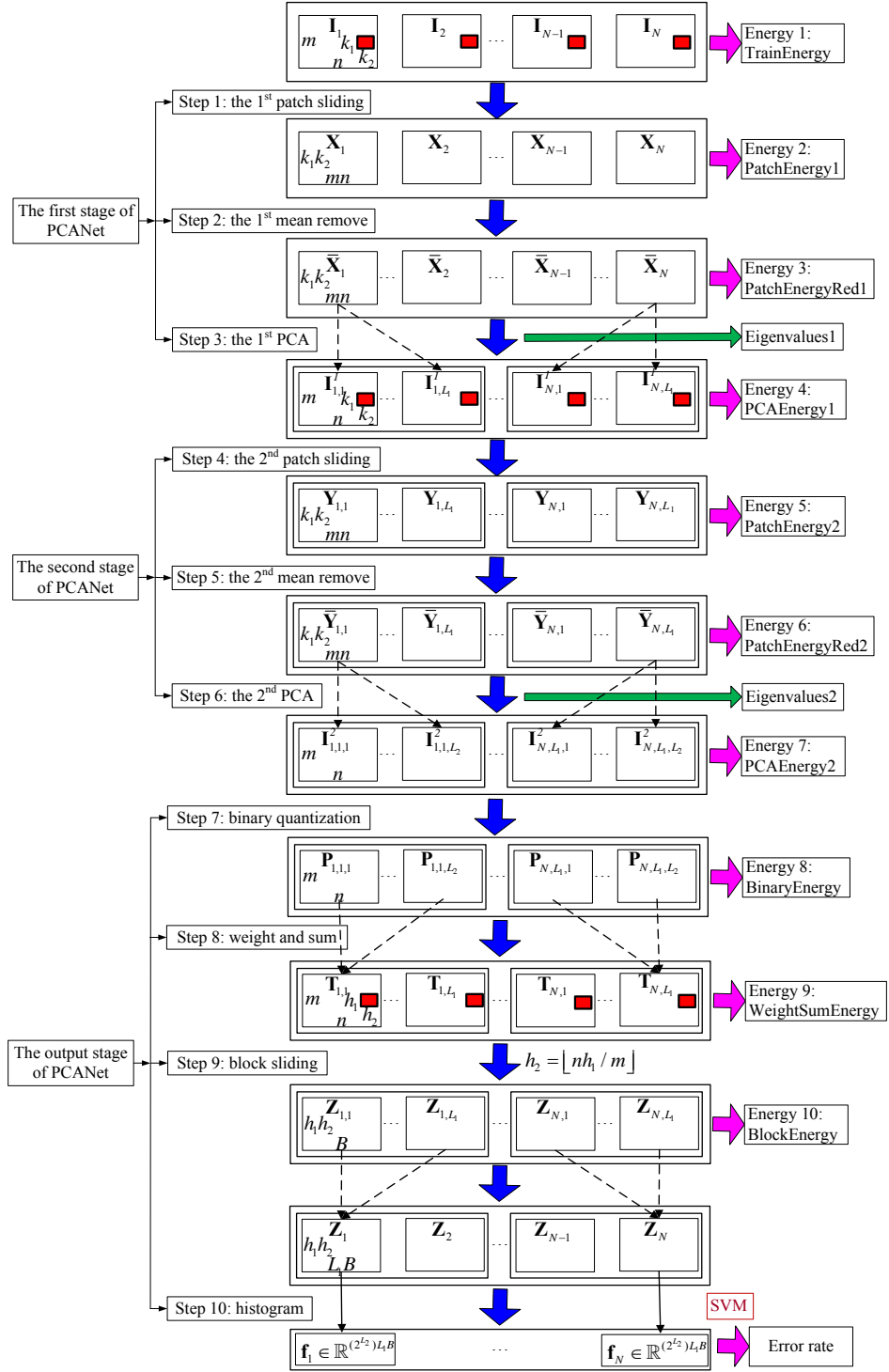

Fig. 1. The block diagram of PCANet

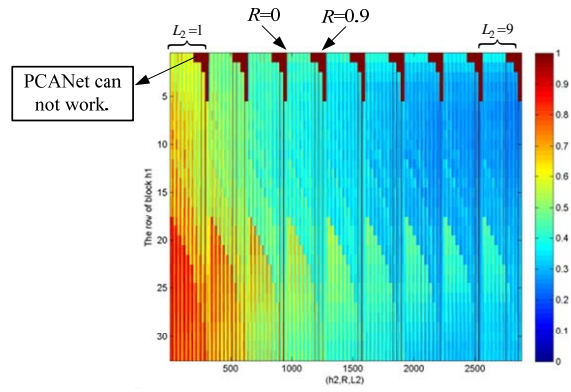

*(a)* $L_1$=1

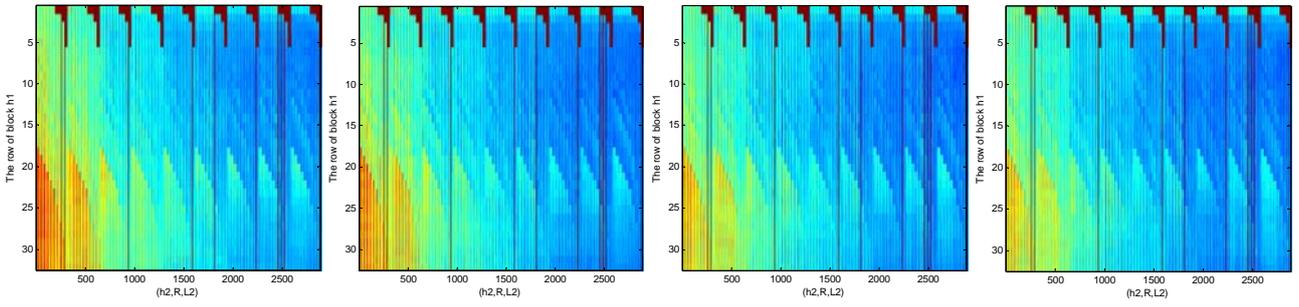

*(b)* $L_1$=2  *(c)* $L_1$=3  *(d)* $L_1$=4  *(e)* $L_1$=5

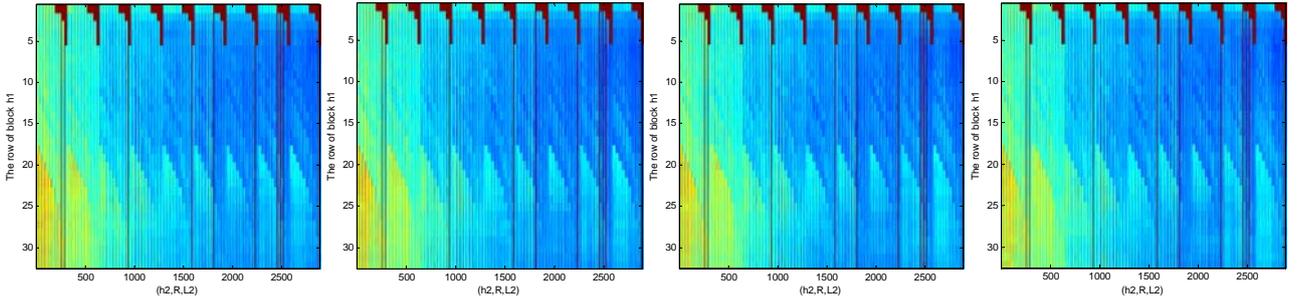

*(f)* $L_1$=6  *(g)* $L_1$=7  *(h)* $L_1$=8  *(i)* $L_1$=9

Fig. 2. The **overalltrend** of error rate *e* with the change of $h_1$, $h_2$, $R$, $L_2$, and $L_1$, where $h_1, h_2 \in \{1:1:32\}$, $L_1, L_2 \in \{1:1:9\}$, and $R \in \{0:0.1:0.9\}$

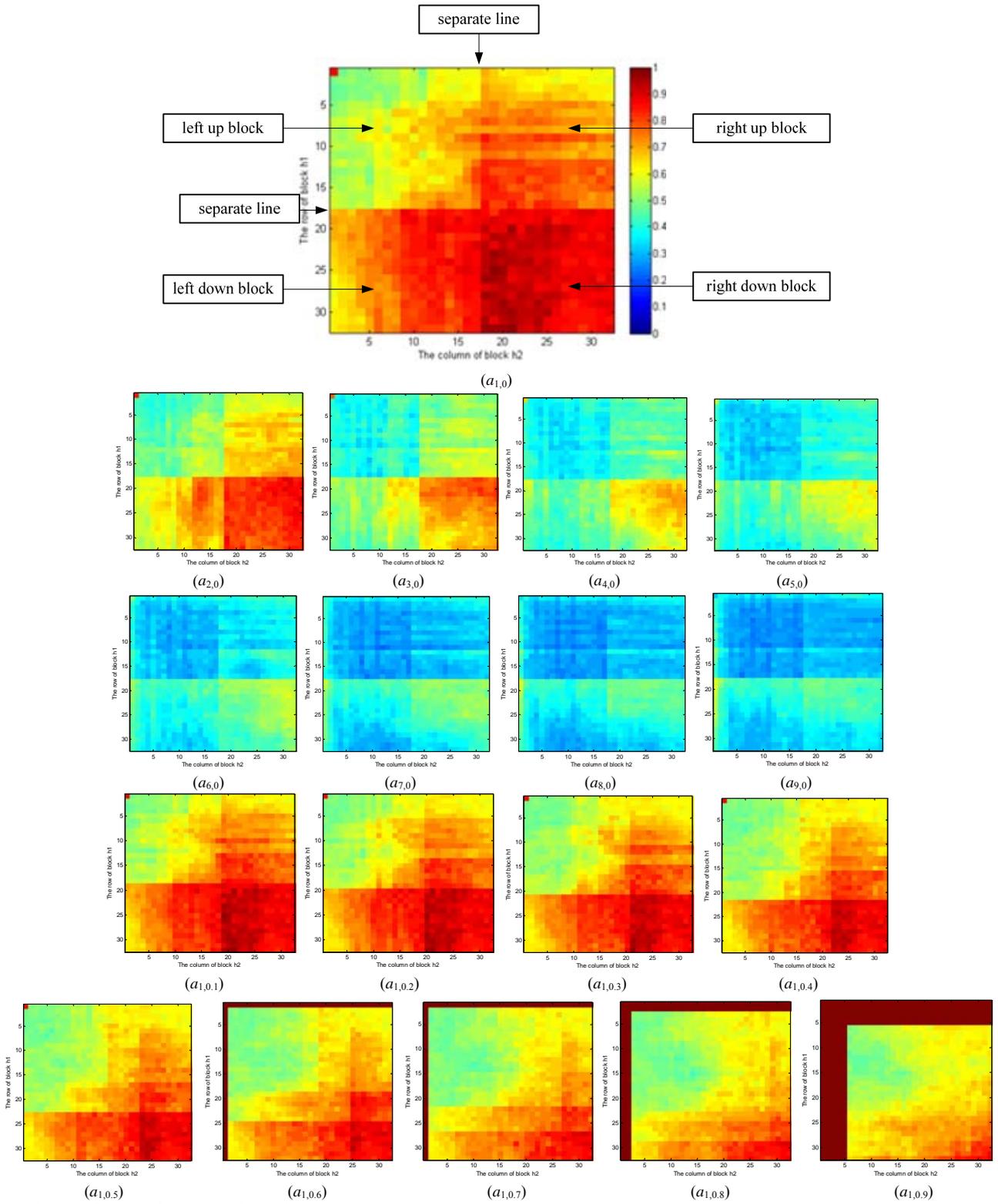

Fig. 3. The part of results of error rate $e$ in relation with $h_1$, $h_2$, $R$ and $L_2$, where $h_1, h_2 \in \{1:1:32\}$, $L_1=1$, $L_2 \in \{1:1:9\}$, and $R \in \{0:0.1:0.9\}$; Note that $a_{1,0}$ means the result of $L_2=1$ and $R=0$.

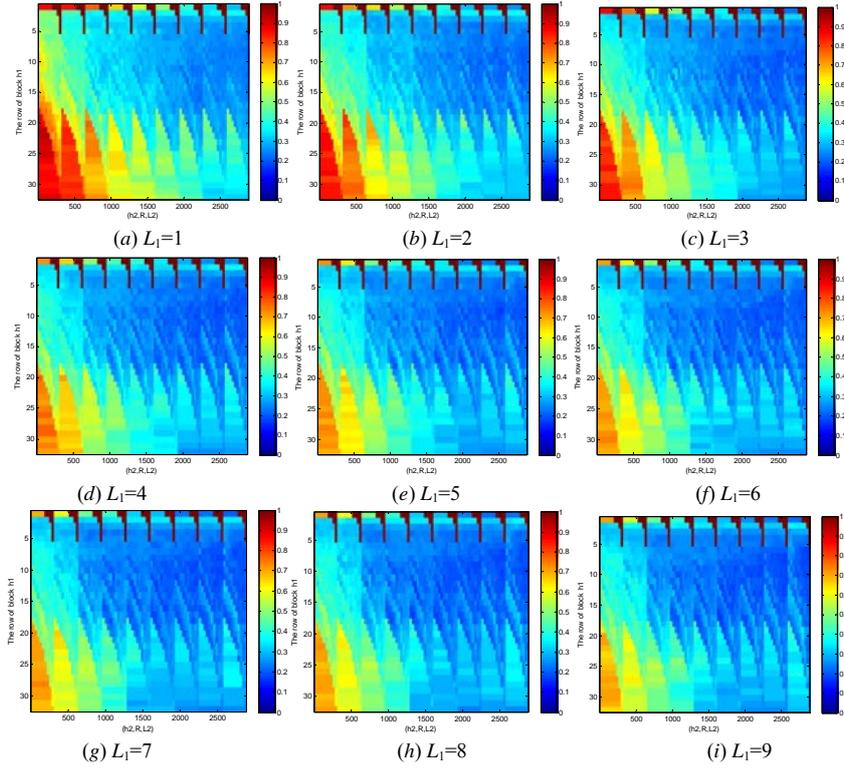

Fig. 4. Error rate $e$ of Yale dataset in relation with $h_1$, $h_2$, $R$, $L_1$, $L_2$, where $h_1 \in \{1:1:32\}$, $h_2 = \lfloor nh_1/m \rfloor$, $R \in \{0:0.1:0.9\}$, $L_1 \in \{1:1:9\}$, $L_2 \in \{1:1:9\}$.

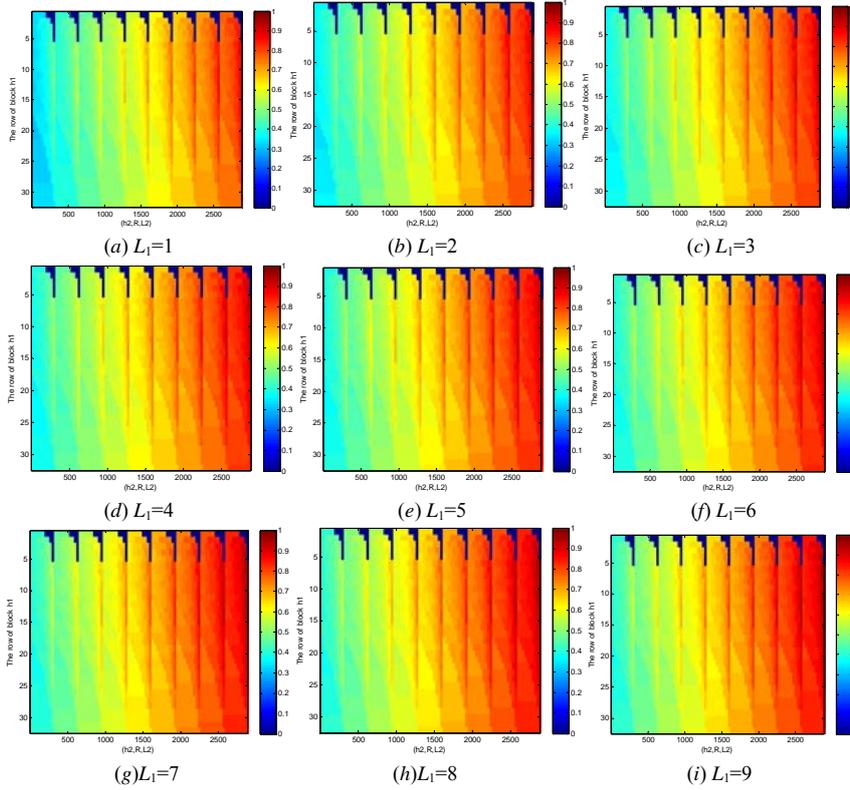

Fig. 5. The logarithm of BlockEnergy of Yale dataset in relation with $h_1$, $h_2$, $R$, $L_1$, $L_2$, where $h_1 \in \{1:1:32\}$, $h_2 = \lfloor nh_1/m \rfloor$, $R \in \{0:0.1:0.9\}$, $L_1 \in \{1:1:9\}$, $L_2 \in \{1:1:9\}$.

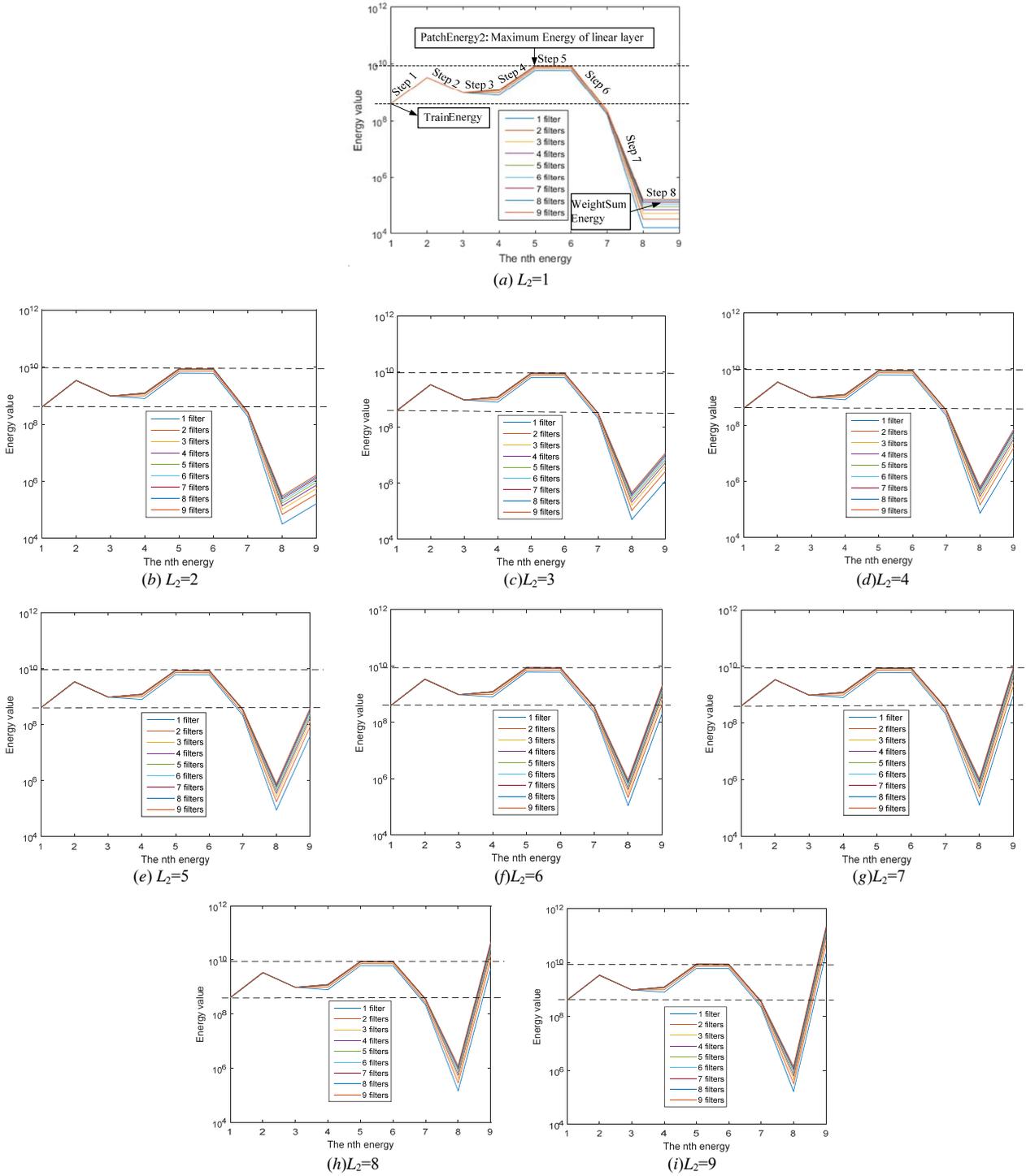

Fig. 6. The energy values of every stages of the first part of PCANet versus the *n*th energy (Yale dataset). The parameters are $L_1, L_2 \in \{1:1:9\}$.

TABLE I THE PARAMETERS OF PCANET.

| Parameters | Descriptions |
|---|---|
| $m \times n$ | The size of input image |
| $k_1 \times k_2$ | The patch size. $k_1$ and $k_2$ are odd integers and satisfy $1 \leq k_1 \leq m$, $1 \leq k_2 \leq n$. In this paper, we always set $k_1=k_2=3$. |
| $L_1, L_2$ | The number of filters of two stages. $1 \leq L_1 \leq k_1 k_2$, $1 \leq L_2 \leq k_1 k_2$ |
| $h_1 \times h_2$ | The block size. $1 \leq h_1 \leq m$, $1 \leq h_2 \leq n$. Constraint: $h_2 = \lfloor n h_1 / m \rfloor$ |
| $R$ | The overlap ratio of block. $R \in \{0:0.1:0.9\}$ which means R varies from 0 to 0.9 with the interval 0.1. |

TABLE II THE ENERGY AND ERROR RATE OF PCANET WE RECORDED AND THEIR DESCRIPTIONS.

| Two parts | Energy or Error rate | Descriptions |
|---|---|---|
| The first part | **TrainEnergy** | The total energy of $N$ input training images. $E(\mathbf{I}) = \sum_{j=1}^{m} \sum_{k=1}^{Nn} \mathbf{I}^2(j,k)$, $\mathbf{I}$ is defined in Eq. (1). |
| | **PatchEnergy1** | The total energy of $N$ images after the first patch sliding process (Step 1). $E(\mathbf{X}) = \sum_{j=1}^{k_1 k_2} \sum_{k=1}^{Nmn} \mathbf{X}^2(j,k)$, $\mathbf{X}$ is defined in Eq. (3). |
| | **PatchEnergyRed 1** | The total energy of $N$ images after the first mean remove process (Step 2). $E(\bar{\mathbf{X}}) = \sum_{j=1}^{k_1 k_2} \sum_{k=1}^{Nmn} \bar{\mathbf{X}}^2(j,k)$, $\bar{\mathbf{X}}$ is defined in Eq. (5). |
| | **PCAEnergy1** | The total energy of $NL_1$ images after the first PCA process (Step 3) $E(\mathbf{I}^1) = \sum_{j=1}^{m} \sum_{k=1}^{L_1 Nn} \left(\mathbf{I}^1(j,k)\right)^2$, $\mathbf{I}^1$ is defined in Eq. (13). |
| | **PatchEnergy2** | The total energy of $NL_1$ images after the second patch sliding process (Step 4). $E(\mathbf{Y}) = \sum_{j=1}^{k_1 k_2} \sum_{k=1}^{L_1 Nmn} \left(\mathbf{Y}(j,k)\right)^2$, $\mathbf{Y}$ is defined in Eq. (16). |
| | **PatchEnergyRed2** | The total energy of $NL_1$ images after the second mean remove process (Step 5). $E(\bar{\mathbf{Y}}) = \sum_{j=1}^{k_1 k_2} \sum_{k=1}^{L_1 Nmn} \left(\bar{\mathbf{Y}}(j,k)\right)^2$, $\bar{\mathbf{Y}}$ is defined in Eq. (17). |
| | **PCAEnergy2** | The total energy of $NL_1 L_2$ images after the second PCA process (Step 6) $E(\mathbf{I}^2) = \sum_{j=1}^{m} \sum_{k=1}^{L_1 L_2 Nn} \left(\mathbf{I}^2(j,k)\right)^2$, $\mathbf{I}^2$ is defined in Eq. (28). |
| | **BinaryEnergy** | The total energy of $NL_1 L_2$ images after the binary process (Step 7) $E(\mathbf{P}) = \sum_{j=1}^{m} \sum_{k=1}^{L_1 L_2 Nn} \left(\mathbf{P}(j,k)\right)^2$, $\mathbf{P}$ is defined in Eq. (30). |
| | **WeightSumEnergy** | The total energy of $NL_1$ images after the weight and sum process (Step 8) $E(\mathbf{T}) = \sum_{j=1}^{m} \sum_{k=1}^{L_1 Nn} \left(\mathbf{T}(j,k)\right)^2$, $\mathbf{T}$ is defined in Eq. (32). |
| The second part | **BlockEnergy** | The total energy of $NL_1$ images after the block sliding process (Step 9). In relation with the parameters $h_1$, $h_2$, and $R$ $E(\mathbf{Z}) = \sum_{j=1}^{m} \sum_{k=1}^{L_1 L_2 Nn} \left(\mathbf{Z}(j,k)\right)^2$, $\mathbf{Z}$ is defined in Eq. (35). |
| | **ErrorRate** | The error rate $e$ of PCANet. |

TABLE III THE FIT OF DATA RESULT AND ITS CRITERIONS FOR FOUR DATABASES

| | Yale database | ARdatabase | CMU PIE database | ORL database |
|---|---|---|---|---|
| Linear model Poly3 for $e = f(g)$, $g = 1/\log(E(\mathbf{Z}))$ | $e = -231.2g^3 + 123.3g^2$ $-17.33g + 0.9932$ | $e = -516.7g^3 + 234.7g^2$ $-26.7g + 0.9956$ | $e = -134.1g^3 + 154.3g^2$ $-23.76g + 0.9998$ | $e = -165g^3 + 119.9g^2$ $-19.02g + 0.9936$ |
| Error sum of squares (SSE) | 188.2 | 696.7 | 518.1 | 327 |
| R-square | 0.7696 | 0.6499 | 0.7123 | 0.735 |
| Rootmeansquareerror (RMSE) | 0.08522 | 0.164 | 0.1414 | 0.1123 |

TABLE IV THE ENERGY AND ENERGY RATIO OF BLOCKENERGY CORRESPONDING TO $R \in \{0:0.1:0.9\}$. THE PARAMETERS ARE $L_1 = L_2 = 6$, $h_1=h_2=8$. (YALE database)

| Overlap ratio $R$ | 0 | 0.1 | 0.2 | 0.3 | 0.4 | 0.5 | 0.6 | 0.7 | 0.8 | 0.9 |
|---|---|---|---|---|---|---|---|---|---|---|
| The $i$th filter | 1st | 2nd | 3rd | 4th | 5th | 6th | 7th | 8th | 9th | 10th |
| The energy of corresponding output image ($\times 10^{10}$) | 1.1296 | 0.3039 | 0.3039 | 0.1270 | 0.0771 | 0.0565 | 0.0394 | 0.0394 | 0.0256 | 0.0255 |
| Energy sum($\times 10^{10}$) | 1.1296 | 1.4334 | 1.7373 | 1.8643 | 1.9415 | 1.9980 | 2.0374 | 2.0769 | 2.1024 | 2.1279 |
| The ratio of energy | 0.5308 | 0.6736 | 0.8165 | 0.8761 | 0.9124 | 0.9390 | 0.9575 | 0.9760 | 0.9880 | 1.0000 |

TABLE V THE ENERGY RATIO AND THE EIGENVALUE RATIO OF 9 FILTERS (THE 1ST AND THE 2ND PCAs). (YALE DATABASE)

| Parameters | The $n$th filter | Energy sum($\times 10^9$) | Energy Ratio | Eigenvalue ($\times 10^4$) | Eigenvalue sum ($\times 10^4$) | Eigenvalue Ratio | Error Rate |
|---|---|---|---|---|---|---|---|
| $h_1=h_2=8$ $R=0.5$ $L_2=1$ $L_1 \in \{1:1:9\}$ | 1 | 0.7984 | 0.6386 | 2.1544 | 2.1544 | 0.6820 | 0.5037 |
| | 2 | 0.9794 | 0.7834 | 0.3884 | 2.5428 | 0.8049 | 0.4519 |
| | 3 | 1.1223 | 0.8977 | 0.3330 | 2.8757 | 0.9103 | 0.4370 |
| | 4 | 1.1785 | 0.9427 | 0.1151 | 2.9908 | 0.9468 | 0.4370 |
| | 5 | 1.2160 | 0.9727 | 0.0789 | 3.0697 | 0.9717 | 0.4148 |
| | 6 | 1.2339 | 0.9870 | 0.0466 | 3.1163 | 0.9865 | 0.4000 |
| | 7 | 1.2410 | 0.9927 | 0.0189 | 3.1352 | 0.9925 | 0.4000 |
| | 8 | 1.2470 | 0.9975 | 0.0152 | 3.1504 | 0.9973 | **0.3926** |
| | 9 | 1.2502 | 1.0000 | 0.0086 | 3.1590 | 1.000 | 0.4074 |
| $h_1=h_2=8$ $R=0.5$ $L_1=8$ $L_2 \in \{1:1:9\}$ | 1 | 0.22807 | 0.6262 | 3.2949 | 3.2949 | 0.6981 | 0.2444 |
| | 2 | 0.26705 | 0.7332 | 0.5776 | 3.8724 | 0.8205 | 0.2000 |
| | 3 | 0.32293 | 0.8867 | 0.4833 | 4.3558 | 0.9229 | 0.2148 |
| | 4 | 0.33850 | 0.9294 | 0.1581 | 4.5138 | 0.9564 | 0.2074 |
| | 5 | 0.35636 | 0.9785 | 0.1115 | 4.6253 | 0.9800 | 0.2148 |
| | 6 | 0.36088 | 0.9909 | 0.0608 | 4.6861 | 0.9929 | 0.2000 |
| | 7 | 0.36222 | 0.9945 | 0.0164 | 4.7026 | 0.9964 | **0.1852** |
| | 8 | 0.36371 | 0.9986 | 0.0137 | 4.7162 | 0.9993 | **0.1926** |
| | 9 | 0.36421 | 1.0000 | 0.0034 | 4.7162 | 1.0000 | 0.2000 |

TABLE VI THE ERROR RATE CORRESPONDS TO $L_1, L_2 \in \{1:1:9\}$ WHEN $H_1=H_2=8$, $R=0.5$. (YALE DATABASE)

| Error rate $L_1$ \ $L_2$ | 1 | 2 | 3 | 4 | 5 | 6 | 7 | 8 | 9 |
|---|---|---|---|---|---|---|---|---|---|
| 1 | 0.5037 | 0.4370 | 0.3259 | 0.3185 | 0.2963 | 0.2667 | 0.2444 | 0.2444 | 0.2444 |
| 2 | 0.4519 | 0.3926 | 0.2963 | 0.3111 | 0.2741 | 0.2370 | 0.2370 | 0.2000 | 0.2074 |
| 3 | 0.4370 | 0.3630 | 0.2963 | 0.2889 | 0.2593 | 0.2519 | 0.2222 | 0.2148 | 0.2000 |
| 4 | 0.4370 | 0.3481 | 0.2741 | 0.2741 | 0.2593 | 0.2444 | 0.2000 | 0.2074 | 0.2000 |
| 5 | 0.4148 | 0.3259 | 0.2741 | 0.2815 | 0.2667 | 0.2148 | 0.2222 | 0.2148 | 0.2000 |
| 6 | 0.4000 | 0.3333 | 0.2815 | 0.2593 | 0.2593 | 0.2222 | 0.2296 | 0.2000 | 0.2074 |
| 7 | 0.4000 | 0.3259 | 0.2741 | 0.2370 | 0.2519 | 0.2148 | 0.2222 | **0.1852** | 0.1926 |
| 8 | 0.3926 | 0.3185 | 0.2593 | 0.2519 | 0.2370 | 0.2074 | 0.2000 | 0.1926 | 0.2000 |
| 9 | 0.4074 | 0.3259 | 0.2741 | 0.2444 | 0.2370 | 0.2148 | 0.2074 | 0.2000 | 0.2000 |

# Supplement document of PCANet: An energy perspective

# List of Figures

# List of Tables

Table S1. The corresponding relationships of various figures and tables of four databases.

|         | Yale Database   | AR Database     | CMU PIE Database | ORL Database    |
|---------|-----------------|-----------------|------------------|-----------------|
| Figures | Fig. S1 (p. 4)  | Fig. S4 (p. 9)  | Fig. S10 (p. 17) | Fig. S14 (p. 20)|
|         | Fig. 2 (p.9)    | Fig. S5 (p. 10) | -                | -               |
|         | Fig. 3 (p.10) or Fig. S2 (p.7) | Fig. S6 (p. 13) | - | - |
|         | Fig. S3 (p.8)   | -               | -                | -               |
|         | Fig. 4 (p.11)   | Fig. S7 (p. 13) | Fig. S11 (p. 18) | Fig. S15 (p. 21)|
|         | Fig. 5 (p.11)   | Fig. S8 (p. 14) | Fig. S12 (p. 18) | Fig. S16 (p. 21)|
|         | Fig. 6 (p.12)   | Fig. S9 (p. 14) | Fig. S13 (p. 19) | Fig. S17 (p. 22)|
| Tables  | Tab. S2 (p. 8)  | Tab. S4 (p. 15) | -                | -               |
|         | Tab. S3 (p. 8)  | Tab. S3 (p. 8)  | Tab. S3 (p. 8)   | Tab. S3 (p. 8)  |
|         | Tab. III (p. 13)| Tab. III (p. 13)| Tab. III (p. 13) | Tab. III (p. 13)|
|         | Tab. IV (p. 14) | Tab. S5 (p. 16) | Tab. S7 (p. 19)  | Tab. S9 (p. 23) |
|         | Tab. V (p. 14)  | Tab. S6 (p. 16) | Tab. S8 (p. 19)  | Tab. S10 (p. 23)|

# 1. The results of Yale Database

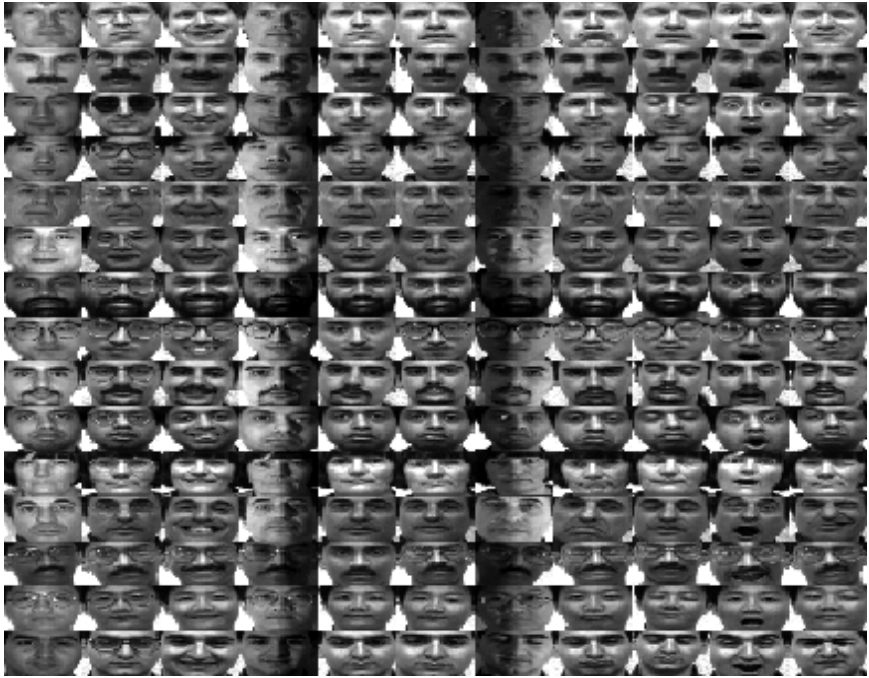

Fig.S1. Images of Yale database

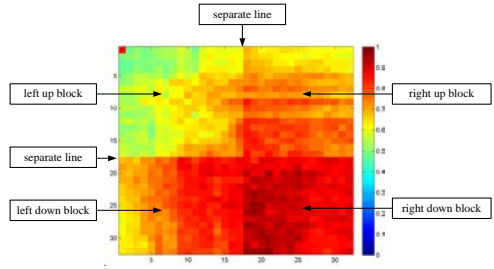

$(a_{1,0})$

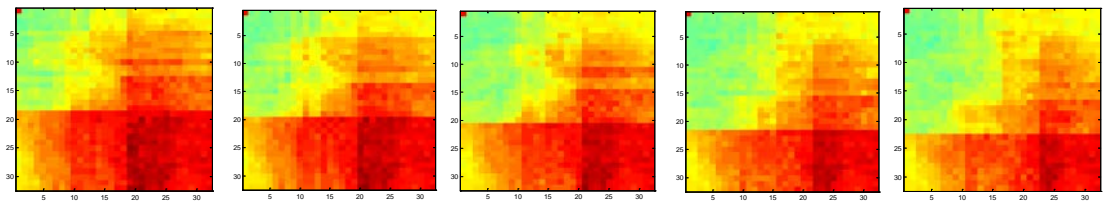

$(a_{1,0.1})$    $(a_{1,0.2})$    $(a_{1,0.3})$    $(a_{1,0.4})$    $(a_{1,0.5})$

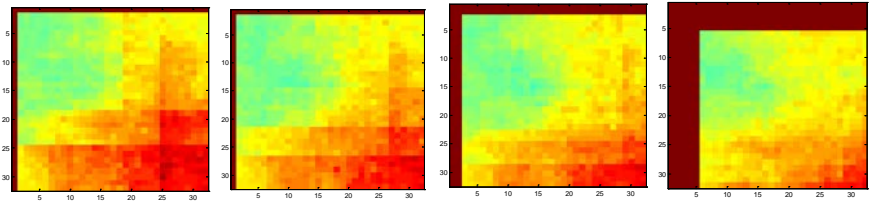

$(a_{1,0.6})$    $(a_{1,0.7})$    $(a_{1,0.8})$    $(a_{1,0.9})$

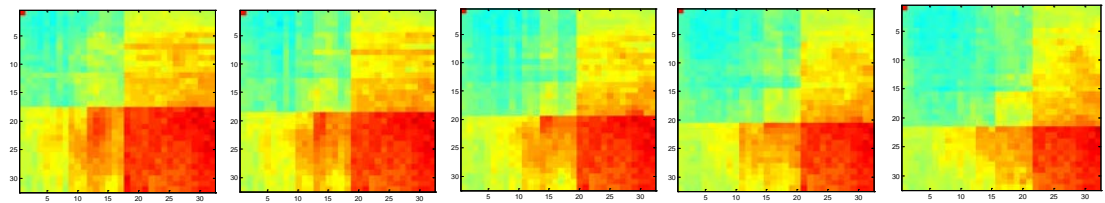

$(a_{2,0})$    $(a_{2,0.1})$    $(a_{2,0.2})$    $(a_{2,0.3})$    $(a_{2,0.4})$

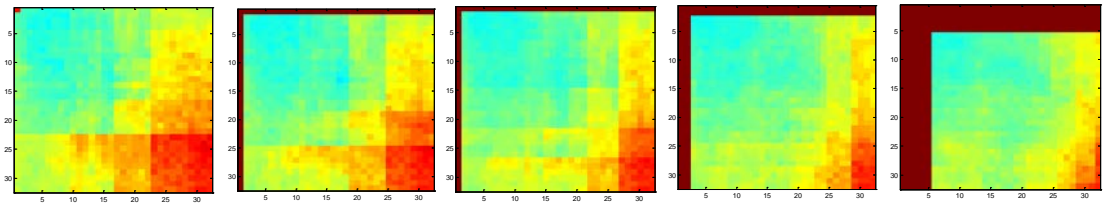

$(a_{2,0.5})$    $(a_{2,0.6})$    $(a_{2,0.7})$    $(a_{2,0.8})$    $(a_{2,0.9})$

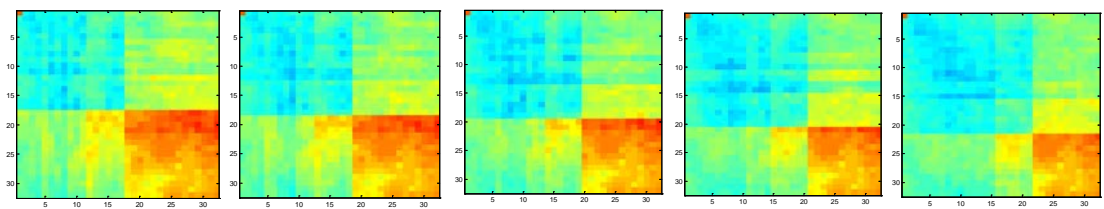

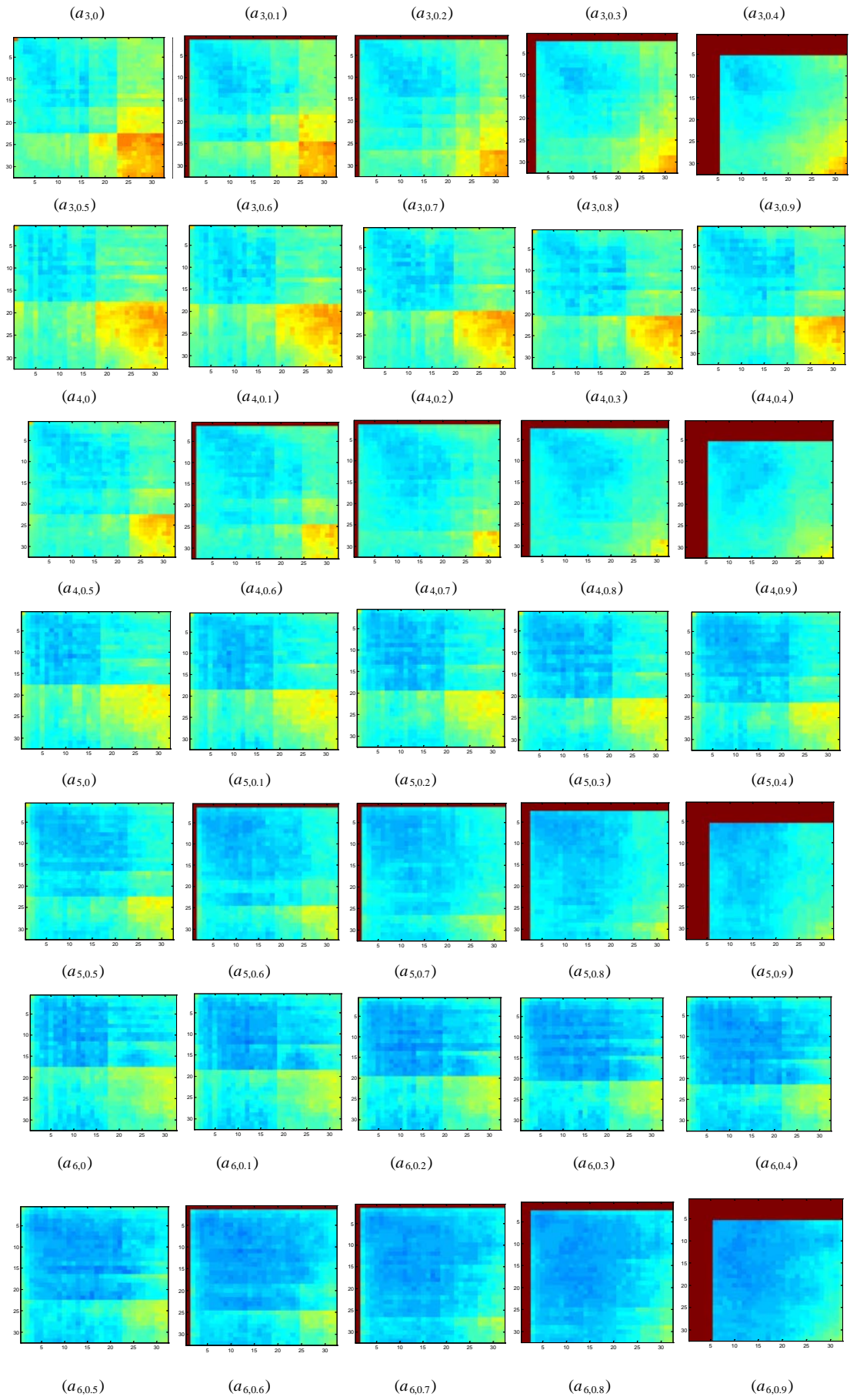

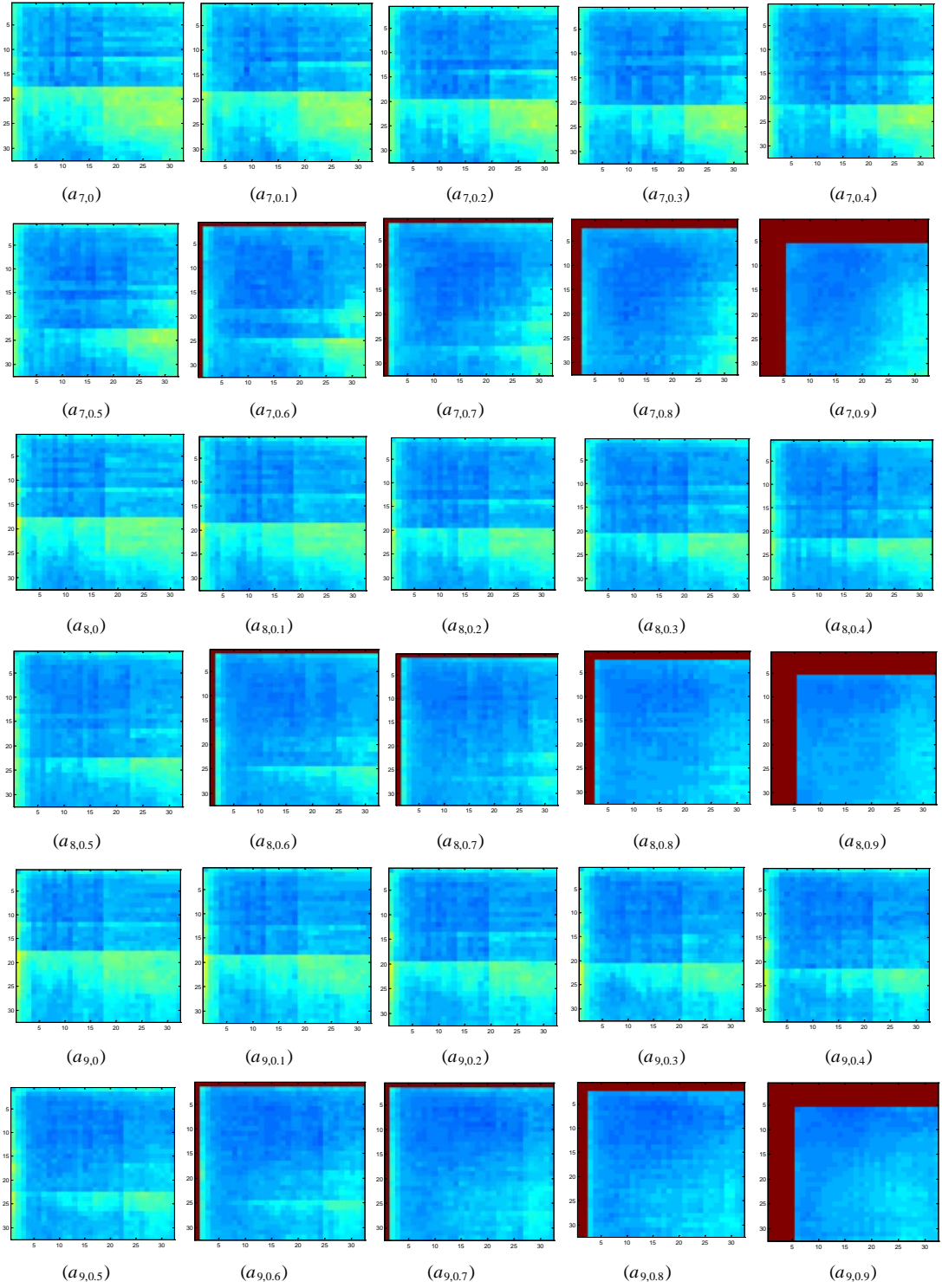

Fig.S2. The results of error rate $e$ in relation with $h_1$, $h_2$, $R$ and $L_2$, where $h_1, h_2 \in \{1:1:32\}$, $L_1=1$, $L_2 \in \{1:1:9\}$, and $R \in \{0:0.1:0.9\}$; Note that $a_{1,0}$ means the result of $L_2=1$ and $R=0$.

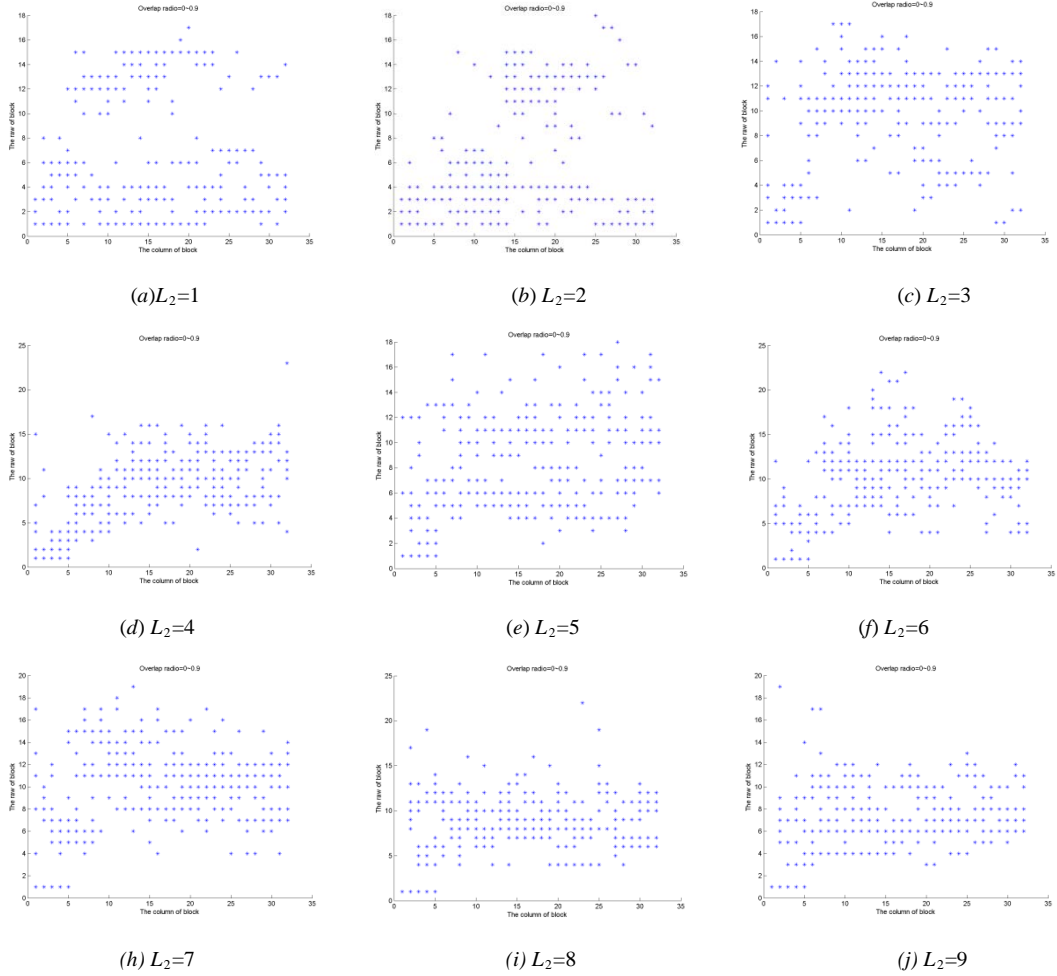

(a) $L_2=1$    (b) $L_2=2$    (c) $L_2=3$

(d) $L_2=4$    (e) $L_2=5$    (f) $L_2=6$

(h) $L_2=7$    (i) $L_2=8$    (j) $L_2=9$

Fig. S3. The value of the row of block $h_1$ that corresponds to the least error rate $e_l$ of all the overlap ratio $R \in \{0:0.1:0.9\}$ when $h_2 \in \{1:1:32\}$, $L_1=1$, $L_2 \in \{1:1:9\}$.

Table S2. The relative error of the least error rate of the diagonal $e_{la}$ and the least error rate of the whole image $e_l$ of Yale database. The parameters are $L_1=1$, $L_2 \in \{1:1:9\}$, $R \in \{0:0.1:0.9\}$, $h_1,h_2 \in \{1:1:32\}$

|   | 0 | 0.1 | 0.2 | 0.3 | 0.4 | 0.5 | 0.6 | 0.7 | 0.8 | 0.9 |
|---|---|---|---|---|---|---|---|---|---|---|
| 1 | 0.0222 | 0.0222 | 0.0296 | 0 | 0.0074 | 0 | 0.0148 | 0.0222 | 0.0296 | 0.0222 |
| 2 | 0.0074 | 0.0296 | 0.0148 | 0 | 0.0074 | 0.0074 | 0.0296 | 0.0074 | 0.0074 | 0.0148 |
| 3 | 0.0222 | 0.0296 | 0.0296 | 0.0074 | 0.0222 | 0.0148 | 0.0074 | 0 | 0.0074 | 0.0074 |
| 4 | 0.0074 | 0.0148 | 0 | 0.0148 | 0.0148 | 0.0074 | 0.0148 | 0 | 0 | 0.0074 |
| 5 | 0 | 0.0222 | 0.0074 | 0.0074 | 0.0222 | 0.0074 | 0.0074 | 0.0148 | 0.0074 | 0.0222 |
| 6 | 0.0074 | 0 | 0.0074 | 0.0074 | 0.0148 | 0.0148 | 0.0370 | 0.0074 | 0.0074 | 0 |
| 7 | 0.0148 | 0.0222 | 0.0222 | 0 | 0.0074 | 0.0074 | 0.0148 | 0.0074 | 0 | 0 |
| 8 | 0 | 0 | 0.0148 | 0.0222 | 0.0074 | 0.0074 | 0.0074 | 0.0222 | 0.0074 | 0.0074 |
| 9 | 0.0074 | 0.0148 | 0.0222 | 0.0074 | 0.0222 | 0.0148 | 0.0148 | 0.0296 | 0.0222 | 0.0222 |

Table S3. The differences of error rates with mean remove step $e_r$ and without mean remove step $e_{wr}$ and their mean values for four databases. The parameters are $L_1=L_2=6$, $R=0.5$, and $1 \leq h_1=h_2 \leq 32$.

| Databases | The values of ($e_r - e_{wr}$) | The mean of ($e_r-e_{wr}$) |
|---|---|---|

| Yale | [-0.0074,0,0,0.0074,0,0,0.0074,0,-0.0074,0,-0.0222,0,0,-0.0074,0.0296,0.0074,0.0148,0,0,0.0370, 0,0,-0.0148,-0.0222,-0.0148,0.0074,-0.0148,0.0148,0.0222,-0.0148,-0.0222,0.0148] | $4.6296 \times 10^{-4}$ |
|---|---|---|
| AR | [-0.0100,-0.0020,0,-0.0020,0.0020,-0.0020,0,0.0040,0.0020,0,-0.0040,0,-0.0020,0.0040,-0.0040,0.0040,-0.0040,0.0060,-0.0040,-0.0060,-0.0040,0.0060,-0.0040,-0.0060,0.0080,0.0060,0.0020,-0.0080,-0.0080, 0.0100, 0, 0.0100] | $-1.8750 \times 10^{-4}$ |
| CMU PIE | [0.0029,0,0,0,0.0004,0,0,0,0,-0.0004,0,0.0004,0.0004,-0.0012,0.0008,-0.0029,-0.0008,0,0.0008, -0.0004,0.0008,-0.0042, 0.0033,-0.0062,0.0071,0.0033,0.0050,0.005,8,0.0033,0.0054,0.0054] | $9.1146 \times 10^{-4}$ |
| ORL | [-0.0062,0.0031,0.0094,0.0031,0,0,0,0.0031,0,0,-0.0125,-0.0031,0.0031,-0.0062,0.0031,-0.0031,0 ,0,0,0.0125,-0.0094,0,-0.0125,-0.0125,-0.0062,-0.0062,-0.0188,-0.0063,-0.0062,0.0031,0,0.0094] | $-19.0 \times 10^{-4}$ |

## 2. The results of AR database

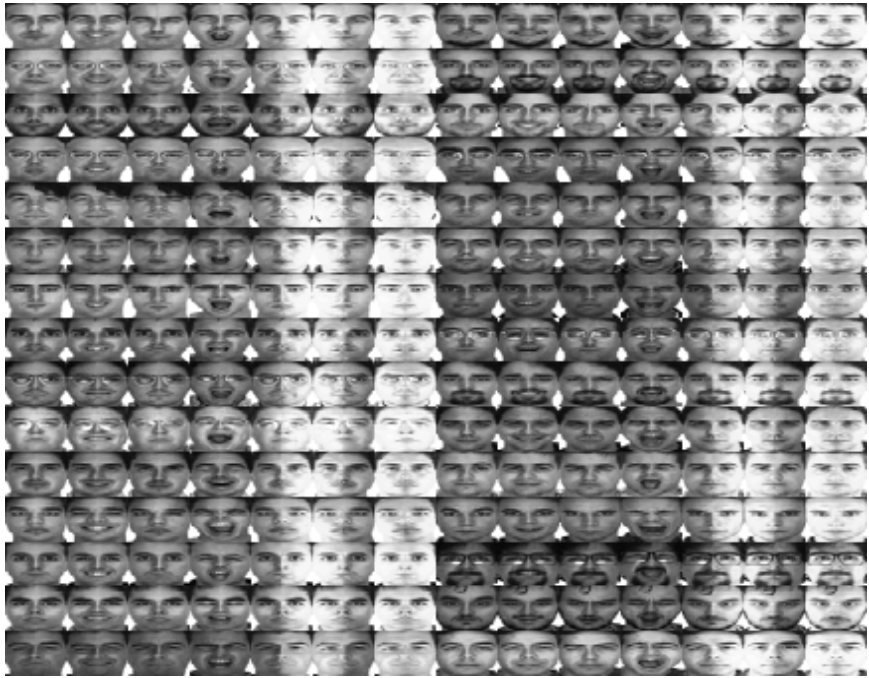

Fig. S4. Images of AR database

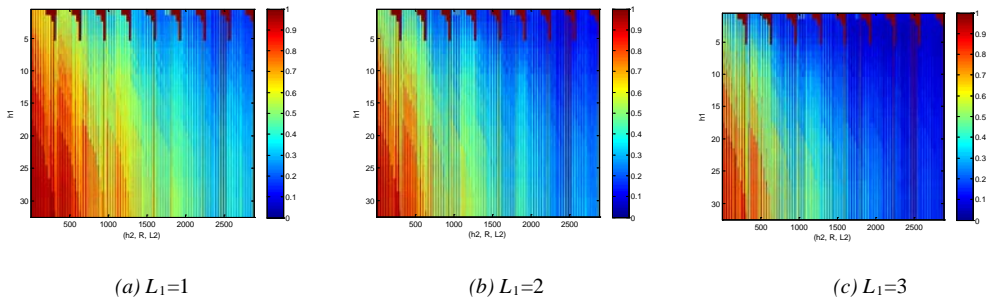

*(a) $L_1$=1*       *(b) $L_1$=2*       *(c) $L_1$=3*

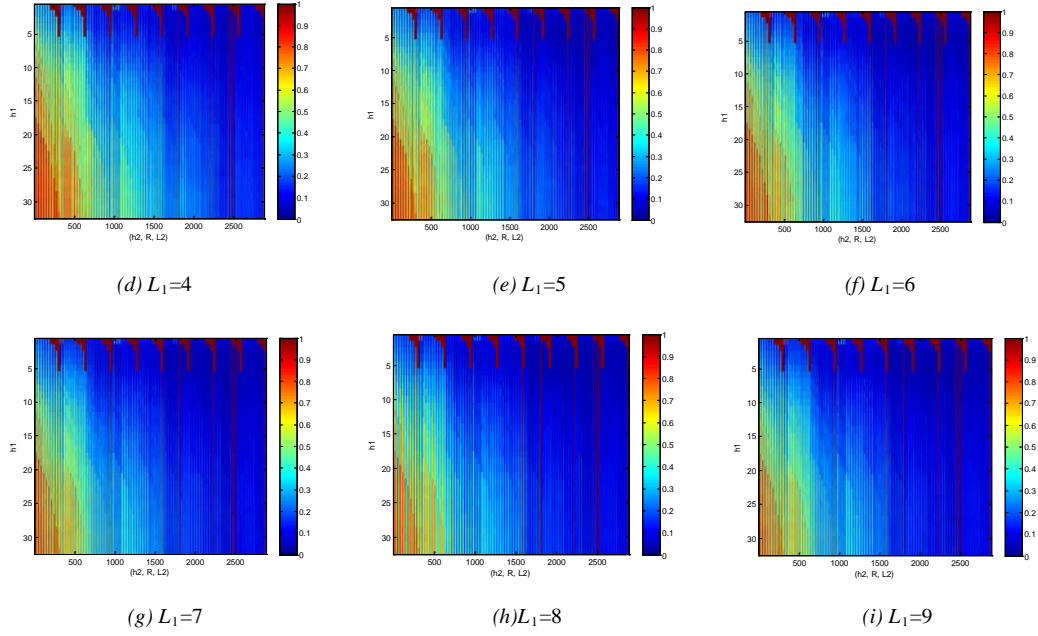

*(d)* $L_1=4$     *(e)* $L_1=5$     *(f)* $L_1=6$

*(g)* $L_1=7$     *(h)* $L_1=8$     *(i)* $L_1=9$

Fig. S5. The **overall trend** of error rate $e$ with the change of $h_1$, $h_2$, $R$, $L_2$, and $L_1$, where $h_1, h_2 \in \{1:1:32\}$, $L_1, L_2 \in \{1:1:9\}$, and $R \in \{0:0.1:0.9\}$. (AR database)

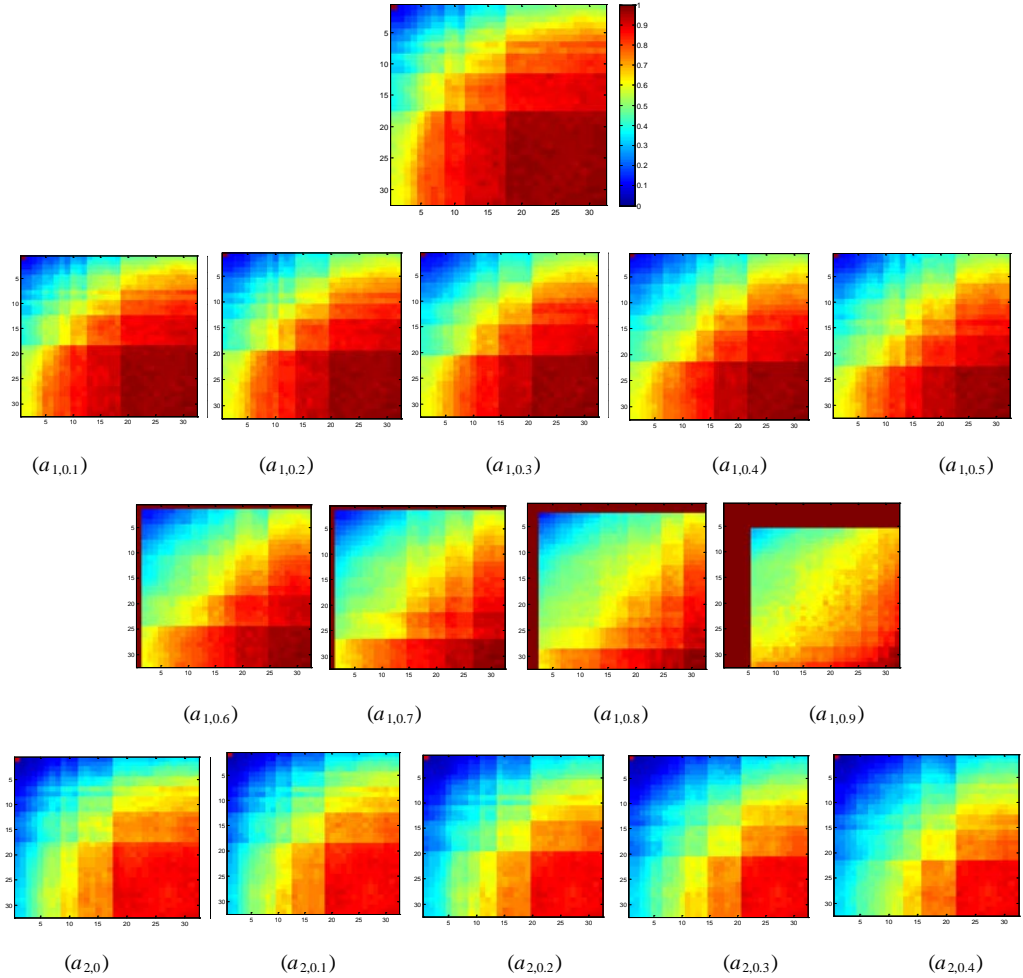

$(a_{1,0.1})$    $(a_{1,0.2})$    $(a_{1,0.3})$    $(a_{1,0.4})$    $(a_{1,0.5})$

$(a_{1,0.6})$    $(a_{1,0.7})$    $(a_{1,0.8})$    $(a_{1,0.9})$

$(a_{2,0})$    $(a_{2,0.1})$    $(a_{2,0.2})$    $(a_{2,0.3})$    $(a_{2,0.4})$

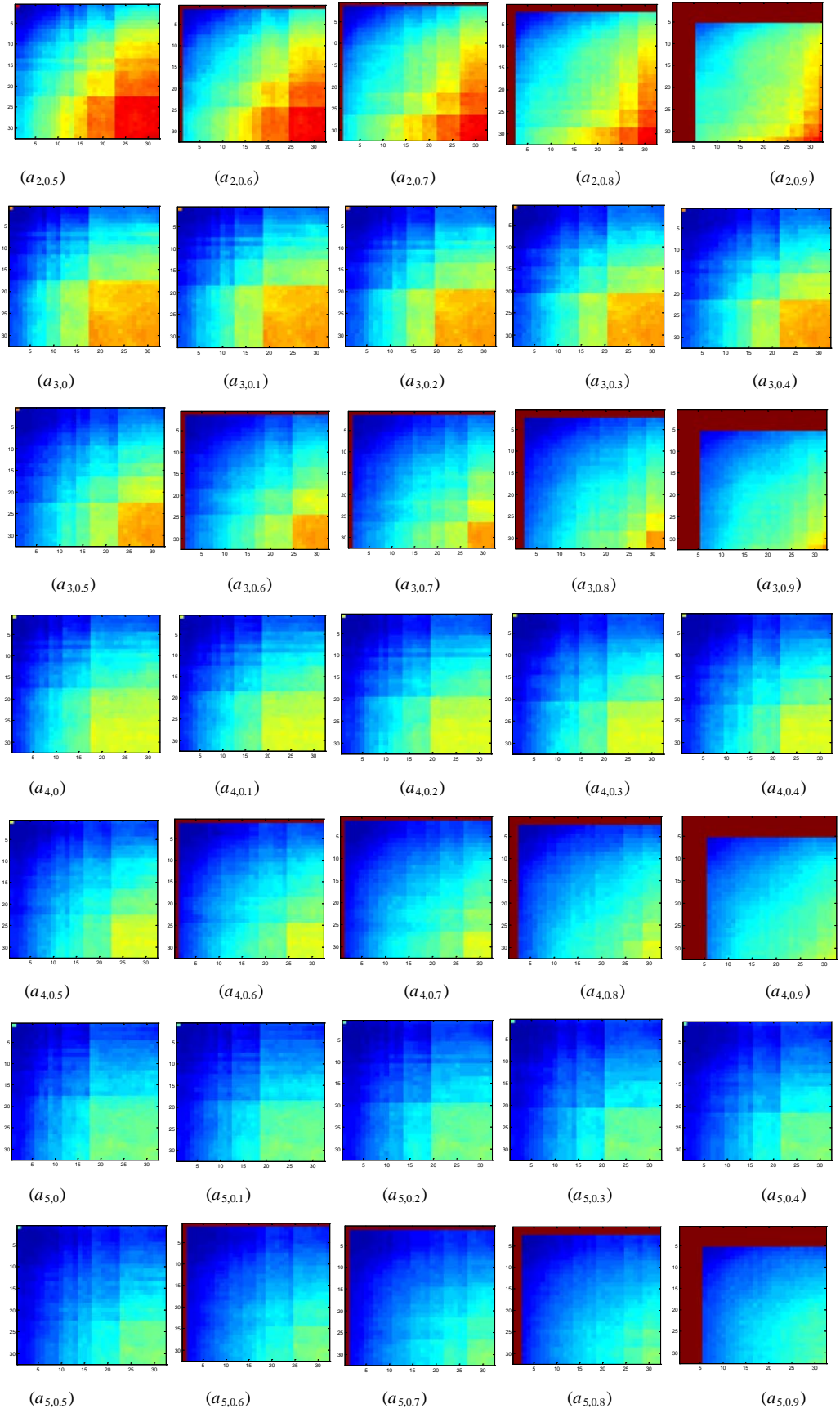

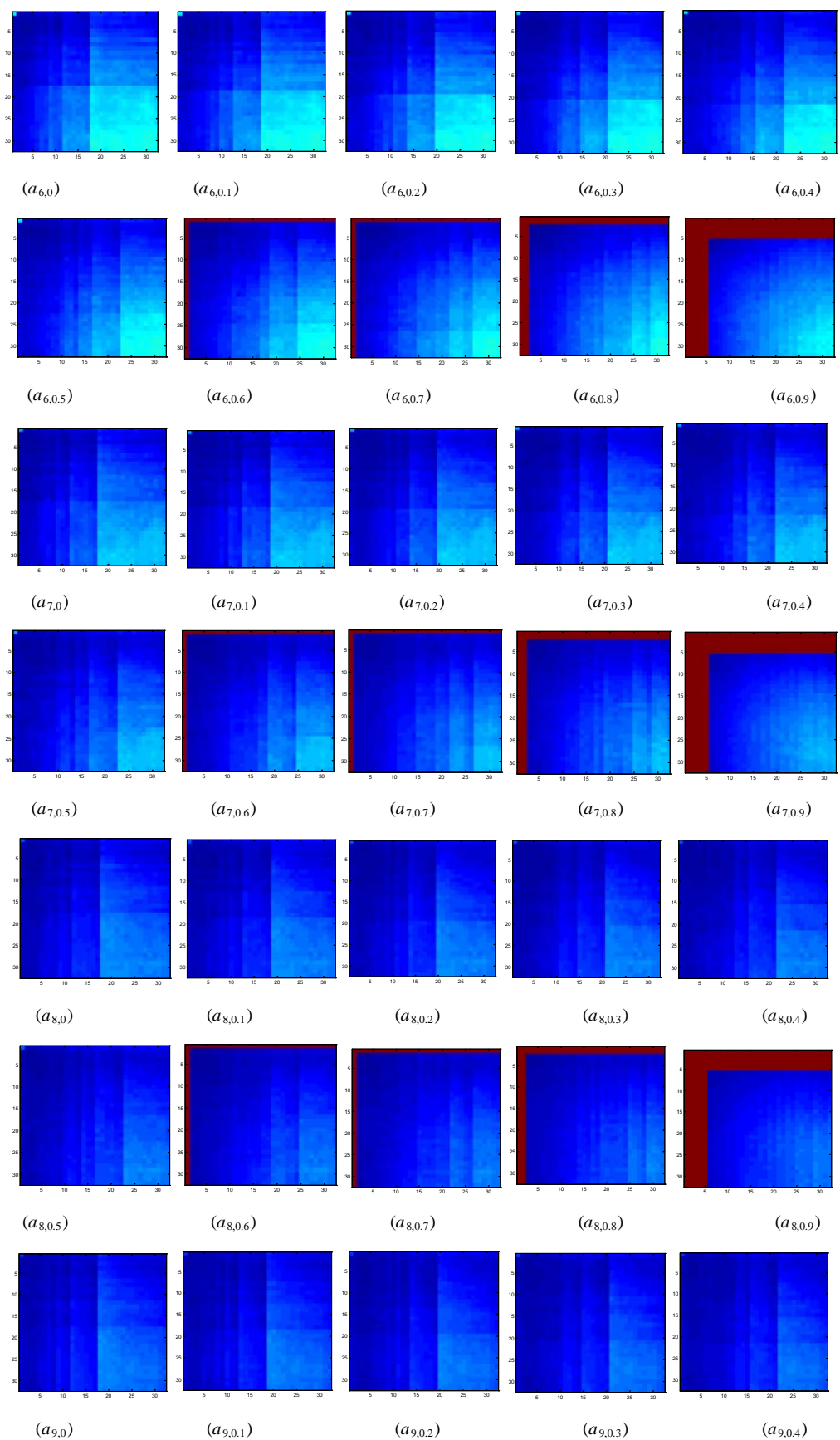

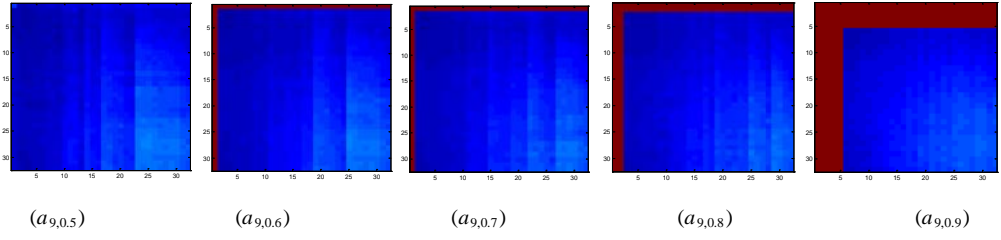

| ($a_{9,0.5}$) | ($a_{9,0.6}$) | ($a_{9,0.7}$) | ($a_{9,0.8}$) | ($a_{9,0.9}$) |

Fig. S6. The results of error rate $e$ in relation with $h_1$, $h_2$, $R$ and $L_2$, where $h_1,h_2 \in \{1:1:32\}$, $L_1=3$, $L_2 \in \{1:1:9\}$, and $R \in \{0:0.1:0.9\}$; Note that $a_{1,0}$ means the result of $L_2=1$ and $R=0$. (AR database)

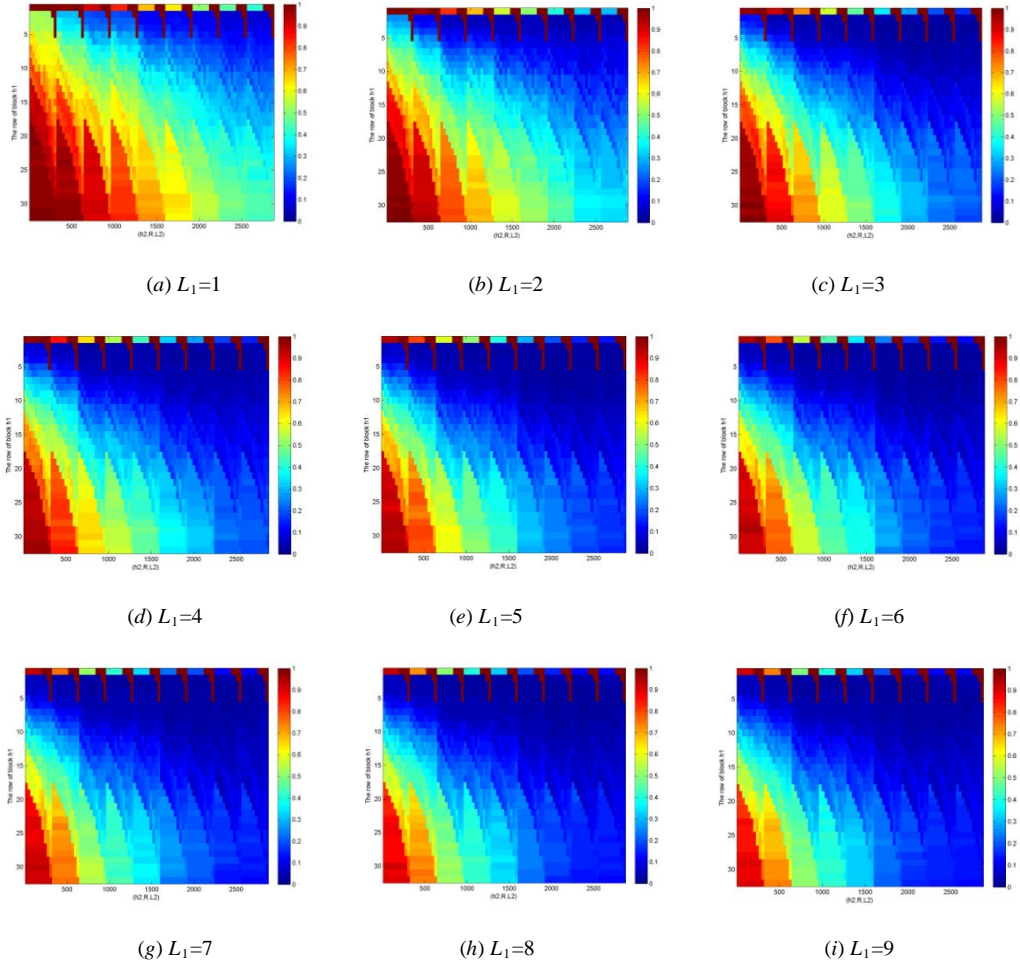

(a) $L_1=1$  (b) $L_1=2$  (c) $L_1=3$

(d) $L_1=4$  (e) $L_1=5$  (f) $L_1=6$

(g) $L_1=7$  (h) $L_1=8$  (i) $L_1=9$

Fig. S7. Error rate $e$ of AR dataset in relation with $h_1$, $h_2$, $R$, $L_1$, $L_2$, where $h_1 \in \{1:1:32\}$, $h_2 = \lfloor nh_1/m \rfloor$, $R \in \{0:0.1:0.9\}$, $L_1 \in \{1:1:9\}$, $L_2 \in \{1:1:9\}$.

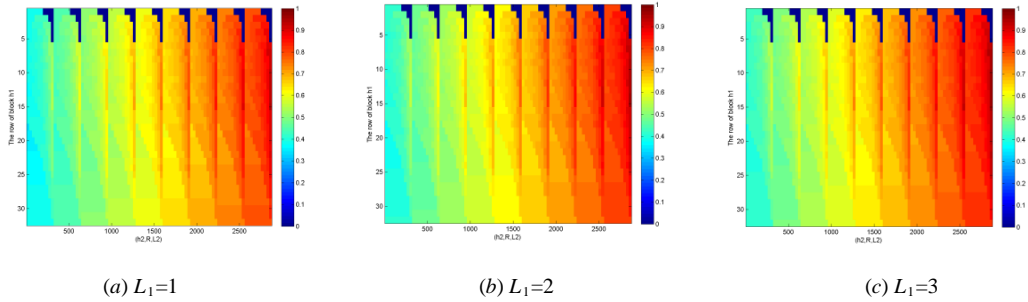

(a) $L_1=1$  (b) $L_1=2$  (c) $L_1=3$

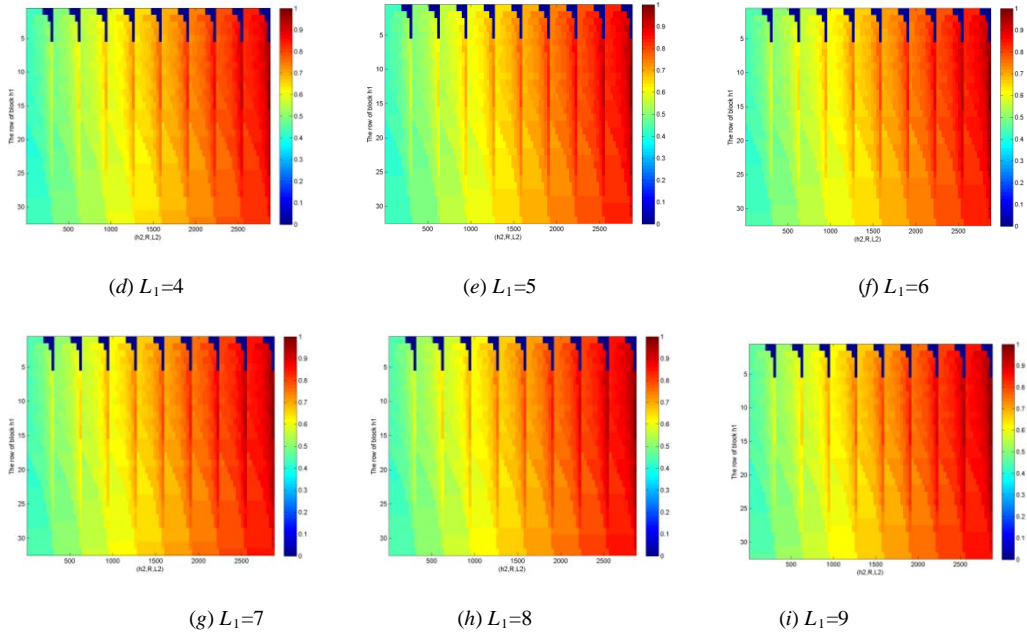

(d) $L_1=4$     (e) $L_1=5$     (f) $L_1=6$

(g) $L_1=7$     (h) $L_1=8$     (i) $L_1=9$

Fig.S8. The logarithm of BlockEnergy of AR dataset in relation with $h_1$, $h_2$, $R$, $L_1$, $L_2$, where $h_1 \in \{1:1:32\}$, $h_2 = \lfloor nh_1/m \rfloor$, $R \in \{0:0.1:0.9\}$, $L_1 \in \{1:1:9\}$, $L_2 \in \{1:1:9\}$.

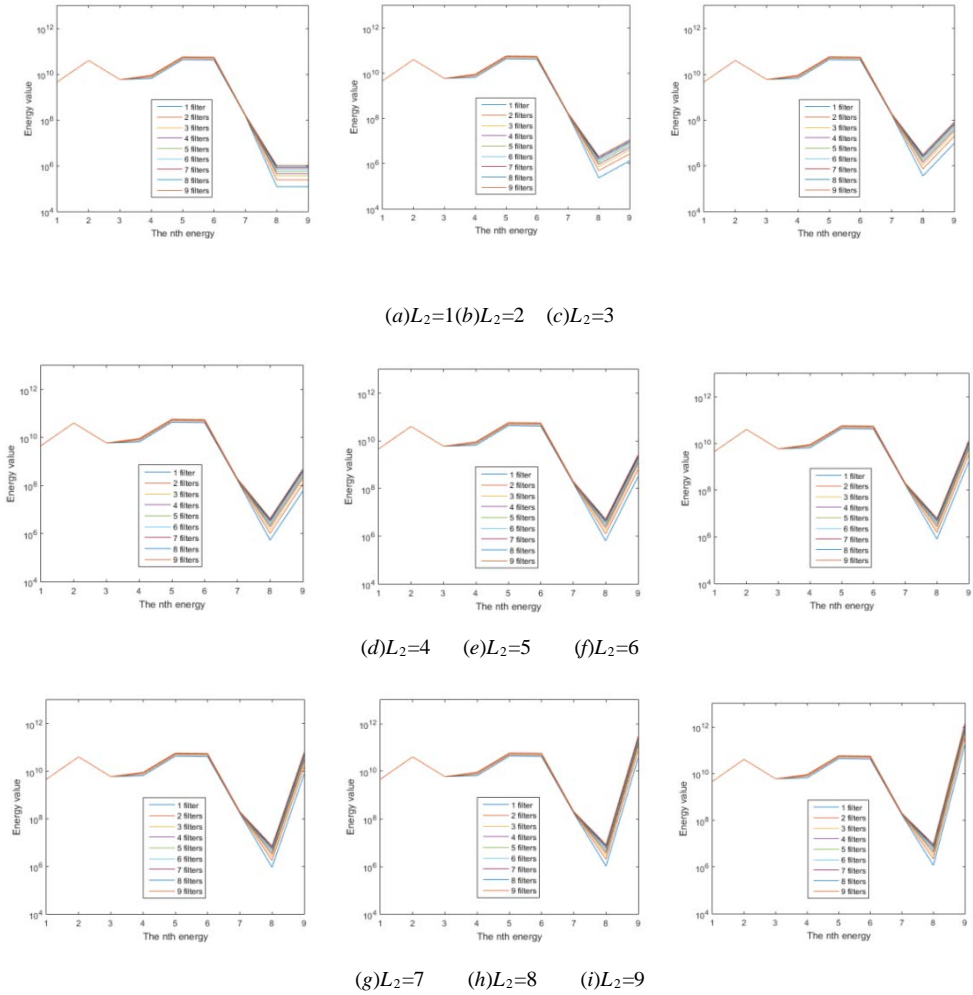

(a) $L_2=1$  (b) $L_2=2$   (c) $L_2=3$

(d) $L_2=4$    (e) $L_2=5$     (f) $L_2=6$

(g) $L_2=7$    (h) $L_2=8$     (i) $L_2=9$

Fig. S9. The energy values of every stages of the first part of PCANet-2 versus the $n$th energy (AR dataset). The parameters are

$L_1$, $L_2 \in \{1:1:9\}$.

Table S4. The relative error of the least error rate of the diagonal $e_{la}$ and the least error rate of the whole image $e_l$ of AR database. The parameters are $L_1$=1, $L_2 \in \{1:1:9\}$, $R \in \{0:0.1:0.9\}$, $h_1,h_2 \in \{1:1:32\}$. The average relative error is 0.0124.The least average relative error is 0.0074=0.1630-0.1556.

|   | 0 | 0.1 | 0.2 | 0.3 | 0.4 | 0.5 | 0.6 | 0.7 | 0.8 | 0.9 |
|---|---|---|---|---|---|---|---|---|---|---|
| 1 | 0 | 0 | 0 | 0 | 0.0060 | 0.0060 | 0.0060 | 0.0040 | 0.0040 | 0.0100 |
| 2 | 0.0060 | 0.0060 | 0.0060 | 0 | 0 | 0 | 0 | 0 | 0 | 0 |
| 3 | 0 | 0 | 0.0040 | 0.0040 | 0.0040 | 0.0040 | 0 | 0 | 0 | 0 |
| 4 | 0.0140 | 0.0140 | 0.0140 | 0 | 0 | 0 | 0 | 0 | 0 | 0 |
| 5 | 0 | 0 | 0 | 0 | 0 | 0 | 0 | 0 | 0 | 0 |
| 6 | 0 | 0 | 0 | 0 | 0 | 0 | 0 | 0 | 0 | 0 |
| 7 | 0.0120 | 0.0120 | 0.0060 | 0 | 0 | 0 | 0.0020 | 0.0020 | 0 | 0 |
| 8 | 0.0060 | 0.0060 | 0 | 0 | 0 | 0 | 0 | 0 | 0 | 0.0020 |
| 9 | 0 | 0 | 0.0080 | 0.0040 | 0 | 0 | 0 | 0 | 0 | 0 |

Table S5. The energy and energy ratio of BlockEnergy corresponding to $R \in \{0:0.1:0.9\}$. The parameters are $L_1 = L_2 = 6$, $h_1=h_2=8$. (AR dataset)

| Overlap ratio $R$ | 0 | 0.1 | 0.2 | 0.3 | 0.4 | 0.5 | 0.6 | 0.7 | 0.8 | 0.9 |
|---|---|---|---|---|---|---|---|---|---|---|
| The $i$th filter | 1st | 2nd | 3rd | 4th | 5th | 6th | 7th | 8th | 9th | 10th |
| The energy of corresponding output image ($\times 10^{10}$) | 7.8248 | 2.0974 | 2.0974 | 0.8832 | 0.5337 | 0.3882 | 0.2713 | 0.2713 | 0.1772 | 0.1738 |
| Energy sum ($\times 10^{11}$) | 0.7825 | 0.9922 | 1.2020 | 1.2903 | 1.3436 | 1.3825 | 1.4096 | 1.4367 | 1.4544 | 1.4718 |
| The ratio of energy | 0.5316 | 0.6741 | 0.8166 | 0.8766 | 0.9129 | 0.9393 | 0.9577 | 0.9762 | 0.9882 | 1.0000 |

Table S6. The energy ratio and the eigenvalue ratio of 9 filters (the 1st and the 2nd PCAs). (AR dataset)

| Parameters | The $n$th filter | Energy values ($\times 10^9$) | Energy Ratio | Eigenvalue ($\times 10^4$) | Eigenvalue sum ($\times 10^4$) | Eigenvalue Ratio | Error Rate |
|---|---|---|---|---|---|---|---|
| $h_1=h_2=8$ $R=0.5$ $L_2=1$ $L_1 \in \{1:1:9\}$ | 1 | 6.5727 | 0.7202 | 2.0927 | 2.0927 | 0.7192 | 0.6120 |
| | 2 | 7.6048 | 0.8333 | 0.3222 | 2.4149 | 0.8299 | 0.4840 |
| | 3 | 8.4639 | 0.9274 | 0.3035 | 2.7184 | 0.9342 | 0.4000 |
| | 4 | 8.7505 | 0.9588 | 0.0766 | 2.7950 | 0.9605 | 0.3400 |
| | 5 | 8.9863 | 0.9847 | 0.0616 | 2.8567 | 0.9817 | 0.2800 |
| | 6 | 9.0687 | 0.9937 | 0.0318 | 2.8884 | 0.9926 | 0.2500 |
| | 7 | 9.0940 | 0.9965 | 0.0094 | 2.8978 | 0.9958 | 0.2280 |
| | 8 | 9.1179 | 0.9991 | 0.0090 | 2.9068 | 0.9989 | 0.2140 |
| | 9 | 9.1264 | 1.0000 | 0.0031 | 2.9100 | 1.0000 | 0.1980 |
| $h_1=h_2=8$ $R=0.5$ $L_1=9$, $L_2 \in \{1:1:9\}$ | 1 | 0.16808 | 0.8217 | 3.4252 | 3.4252 | 0.7476 | 0.1980 |
| | 2 | 0.18053 | 0.8826 | 0.4644 | 3.8896 | 0.8489 | 0.1280 |
| | 3 | 0.19529 | 0.9548 | 0.4237 | 4.3133 | 0.9414 | 0.0460 |
| | 4 | 0.19835 | 0.9697 | 0.1094 | 4.4227 | 0.9653 | 0.0420 |
| | 5 | 0.20276 | 0.9913 | 0.0972 | 4.5199 | 0.9865 | 0.0460 |
| | 6 | 0.20391 | 0.9969 | 0.0422 | 4.5621 | 0.9957 | 0.0300 |
| | 7 | 0.20422 | 0.9984 | 0.0092 | 4.5713 | 0.9977 | 0.0200 |
| | 8 | 0.20446 | 0.9996 | 0.0089 | 4.5802 | 0.9996 | 0.0260 |
| | 9 | 0.20454 | 1.0000 | 0.0016 | 4.5818 | 1.0000 | 0.0280 |

# 3. The results of CMU PIE Database

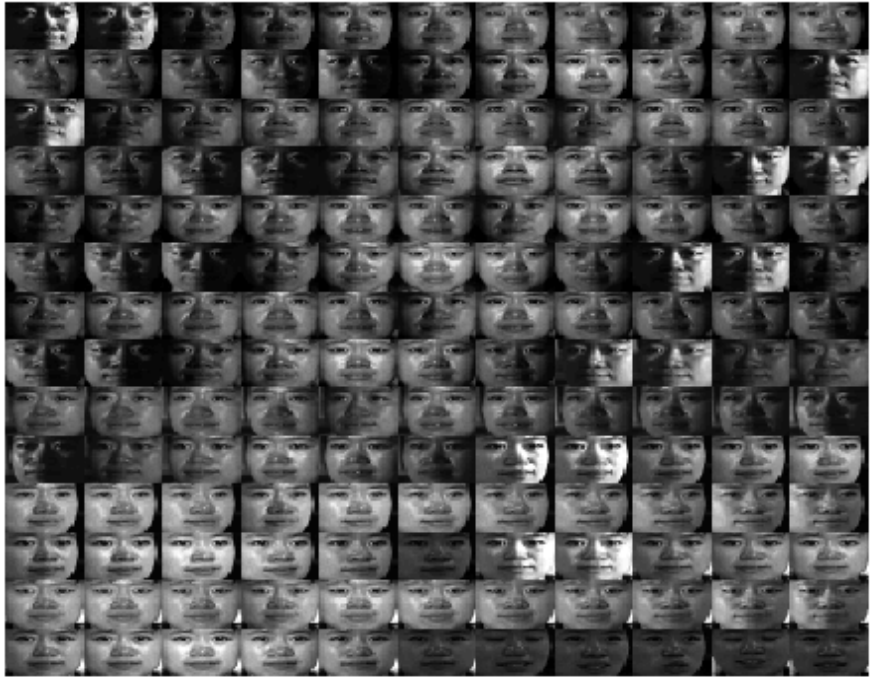

Fig. S10. Some images of CMU PIE database

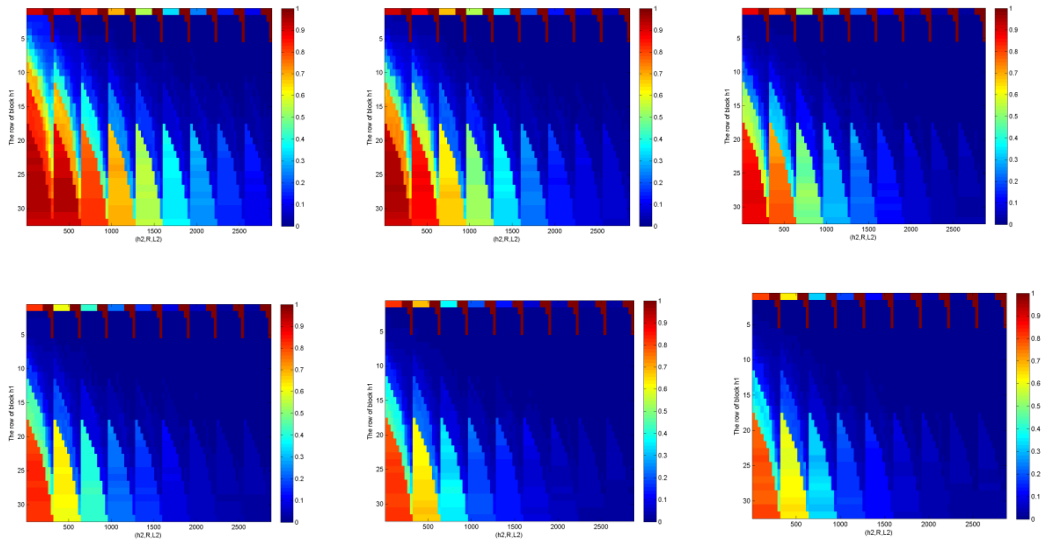

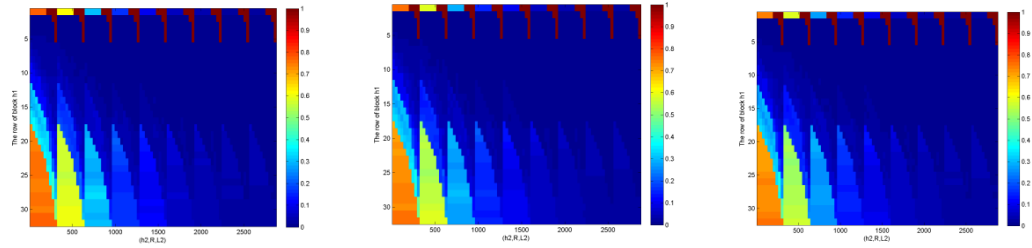

Fig.S11. Error rate $e$ of CMU PIE dataset in relation with $h_1$, $h_2$, $R$, $L_1$, $L_2$, where $h_1 \in \{1:1:32\}$, $h_2 = \lfloor nh_1/m \rfloor$, $R \in \{0:0.1:0.9\}$, $L_1 \in \{1:1:9\}$, $L_2 \in \{1:1:9\}$.

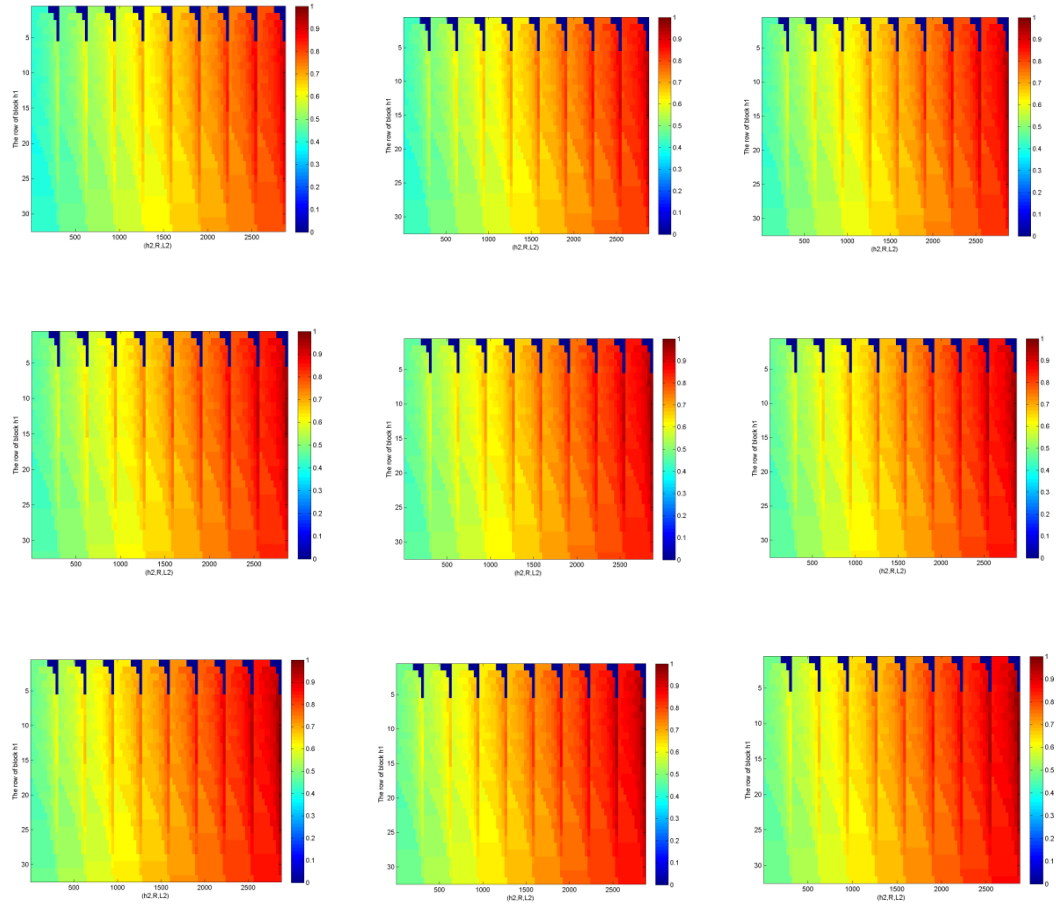

Fig. S12. The logarithm of BlockEnergy of CMU PIE dataset in relation with $h_1$, $h_2$, $R$, $L_1$, $L_2$, where $h_1 \in \{1:1:32\}$, $h_2 = \lfloor nh_1/m \rfloor$, $R \in \{0:0.1:0.9\}$, $L_1 \in \{1:1:9\}$, $L_2 \in \{1:1:9\}$.

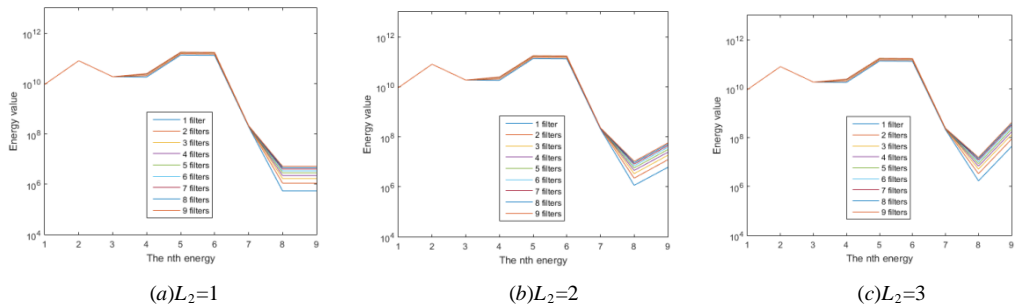

(a) $L_2=1$      (b) $L_2=2$      (c) $L_2=3$

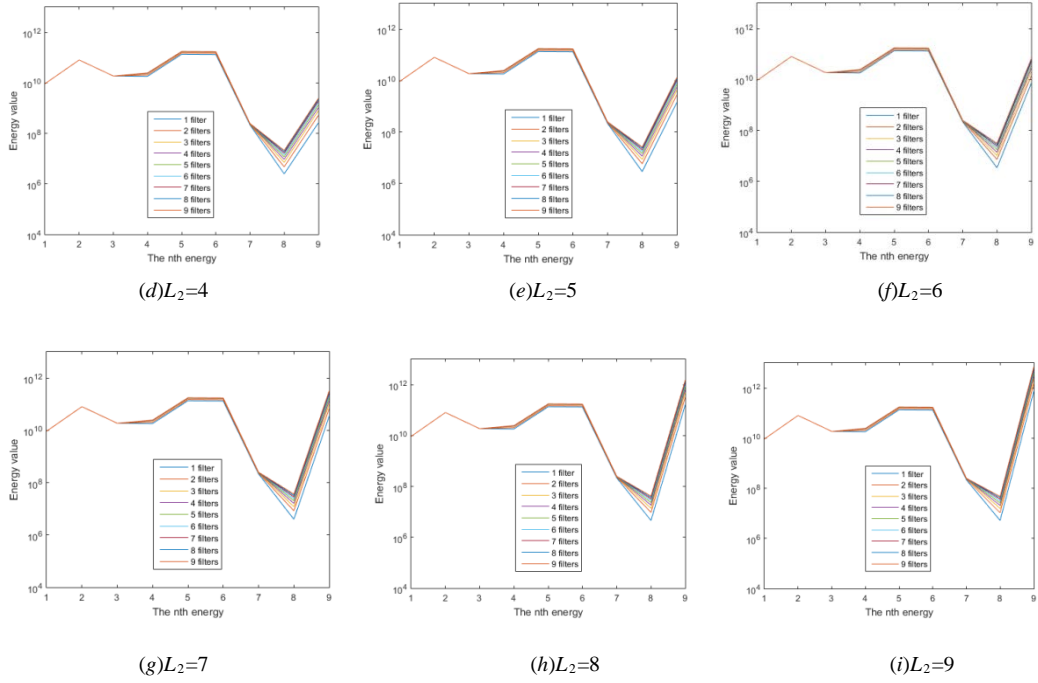

(d)$L_2=4$    (e)$L_2=5$    (f)$L_2=6$

(g)$L_2=7$    (h)$L_2=8$    (i)$L_2=9$

Fig. S13. The energy values of every stages of the first part of PCANet-2 versus the $n$th energy (CMU PIE dataset). The parameters are $L_1$, $L_2 \in \{1:1:9\}$.

Table S7. The energy and energy ratio of BlockEnergy corresponding to $R \in \{0:0.1:0.9\}$. The parameters are $L_1 = L_2 = 6$, $h_1=h_2=8$. (CMU PIE dataset)

| Overlap ratio $R$ | 0 | 0.1 | 0.2 | 0.3 | 0.4 | 0.5 | 0.6 | 0.7 | 0.8 | 0.9 |
|---|---|---|---|---|---|---|---|---|---|---|
| The $i$th filter | 1st | 2nd | 3rd | 4th | 5th | 6th | 7th | 8th | 9th | 10th |
| The energy of corresponding output image ($\times 10^{11}$) | 3.7665 | 1.0084 | 1.0084 | 0.4167 | 0.2519 | 0.1843 | 0.1275 | 0.1275 | 0.0815 | 0.0808 |
| Energy sum($\times 10^{11}$) | 3.7665 | 4.7749 | 5.7833 | 6.2000 | 6.4518 | 6.6361 | 6.7637 | 6.8912 | 6.9727 | 7.0535 |
| The ratio of energy | 0.5340 | 0.6770 | 0.8199 | 0.8790 | 0.9147 | 0.9408 | 0.9589 | 0.9770 | 0.9885 | 1.0000 |

Table S8. The energy ratio and the eigenvalue ratio of 9 filters (the 1st and the 2nd PCAs). (CMU PIE dataset)

| **Parameters** | The $n$th filter | Energy values($\times 10^{10}$) | Energy Ratio | Eigenvalue ($\times 10^4$) | Eigenvalue sum ($\times 10^4$) | Eigenvalue Ratio | Error Rate |
|---|---|---|---|---|---|---|---|
| **$h_1=h_2=8$** **$R=0.5$** **$L_2=1$** **$L_1 \in \{1:1:9\}$** | **1** | **1.8139** | **0.7368** | **1.3782** | **1.3782** | **0.7569** | **0.2471** |
| | **2** | **2.0447** | **0.8306** | **0.1672** | **1.5453** | **0.8487** | **0.0992** |
| | **3** | **2.2886** | **0.9297** | **0.1616** | **1.7069** | **0.9374** | **0.0504** |
| | **4** | **2.3632** | **0.9600** | **0.0429** | **1.7498** | **0.9610** | **0.0392** |
| | **5** | **2.4194** | **0.9828** | **0.0354** | **1.7852** | **0.9804** | **0.0333** |
| | **6** | **2.4422** | **0.9921** | **0.0191** | **1.8043** | **0.9909** | **0.0279** |

| | 7 | 2.4505 | 0.9954 | 0.0070 | 1.8113 | 0.9948 | 0.0250 |
| --- | --- | --- | --- | --- | --- | --- | --- |
| | 8 | 2.4581 | 0.9985 | 0.0064 | 1.8177 | 0.9983 | 0.0238 |
| | 9 | 2.4617 | 1.0000 | 0.0032 | 1.8208 | 1.0000 | 0.0221 |
| $h_1=h_2=8$<br>$R=0.5$<br>$L_1=9$,<br>$L_2\in\{1:1:9\}$ | 1 | 0.021162 | 0.8439 | 2.1506 | 2.1506 | 0.7741 | 0.0221 |
| | 2 | 0.022870 | 0.9120 | 0.2421 | 2.3927 | 0.8612 | 0.0133 |
| | 3 | 0.024114 | 0.9617 | 0.2342 | 2.6269 | 0.9455 | 0.0050 |
| | 4 | 0.024497 | 0.9769 | 0.0633 | 2.6902 | 0.9683 | 0.0033 |
| | 5 | 0.024801 | 0.9890 | 0.0498 | 2.7400 | 0.9862 | 0.0025 |
| | 6 | 0.024957 | 0.9953 | 0.0249 | 2.7649 | 0.9952 | 0.0012 |
| | 7 | 0.025018 | 0.9977 | 0.0063 | 2.7712 | 0.9974 | 0.0017 |
| | 8 | 0.025056 | 0.9992 | 0.0058 | 2.7770 | 0.9995 | 0.0012 |
| | 9 | 0.025076 | 1.0000 | 0.0013 | 2.7784 | 1.0000 | 0.0017 |

## 4. The results of ORL Database

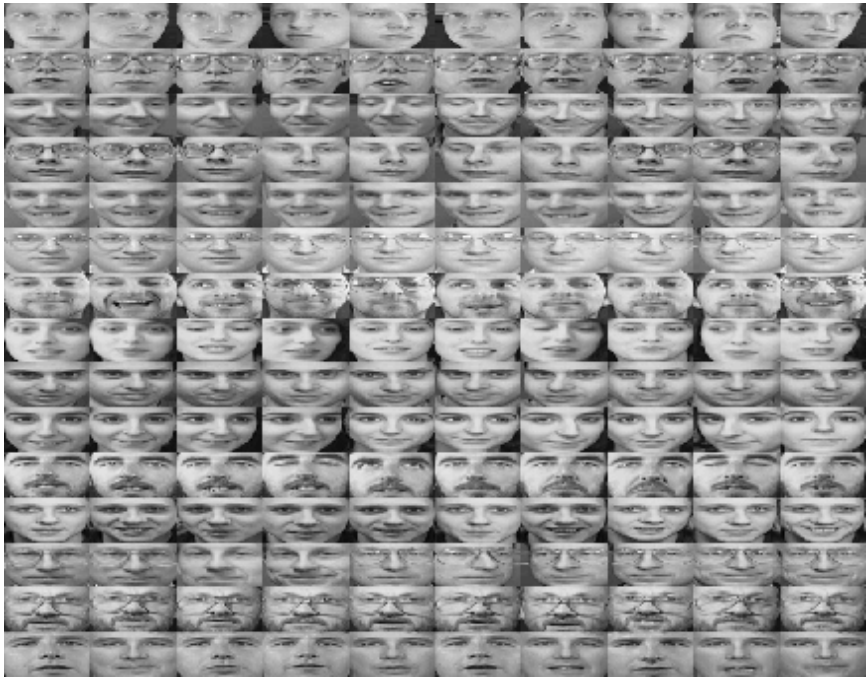

Fig. S14. Some images of ORL database

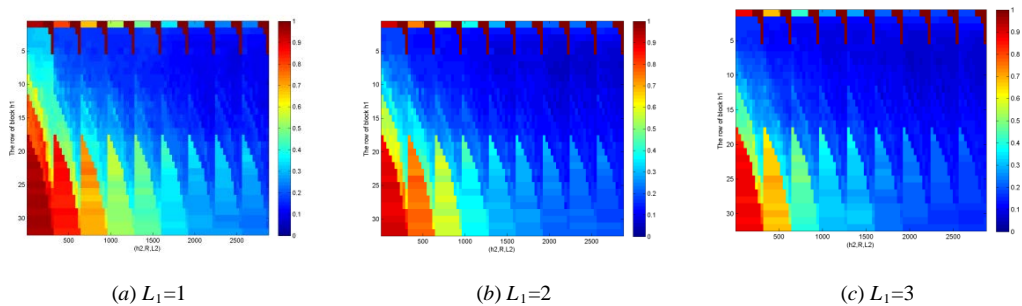

(a) $L_1=1$  (b) $L_1=2$  (c) $L_1=3$

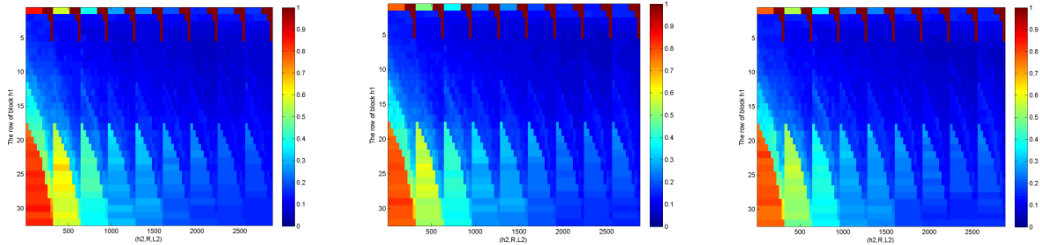

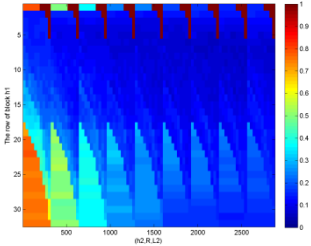
(*d*) $L_1$=4

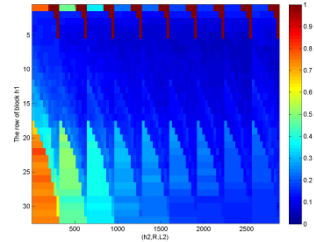
(*e*) $L_1$=5

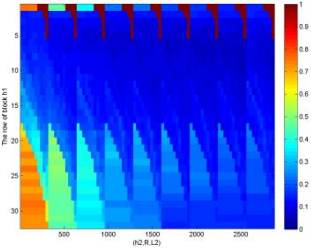
(*f*) $L_1$=6

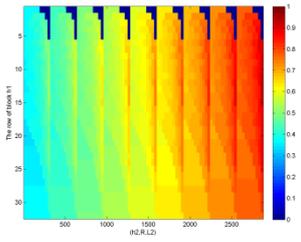
(*g*) $L_1$=7

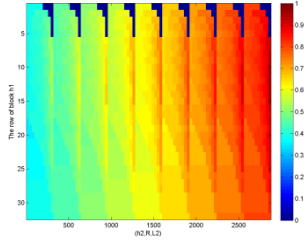
(*h*) $L_1$=8

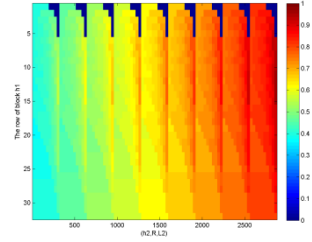
(*i*) $L_1$=9

Fig. S15. Error rate *e* of ORL dataset in relation with $h_1$, $h_2$, $R$, $L_1$, $L_2$, where $h_1 \in \{1:1:32\}$, $h_2 = \lfloor nh_1/m \rfloor$, $R \in \{0:0.1:0.9\}$, $L_1 \in \{1:1:9\}$, $L_2 \in \{1:1:9\}$.

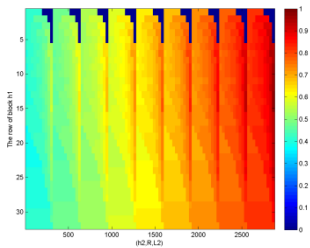
(*a*) $L_1$=1

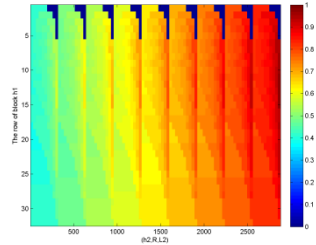
(*b*) $L_1$=2

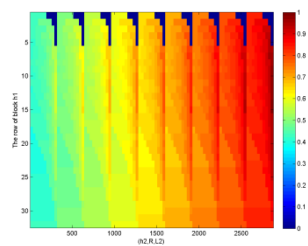
(*c*) $L_1$=3

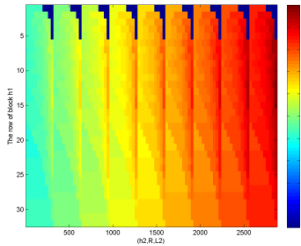
(*d*) $L_1$=4

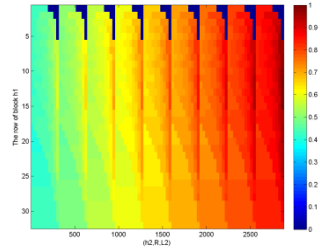
(*e*) $L_1$=5

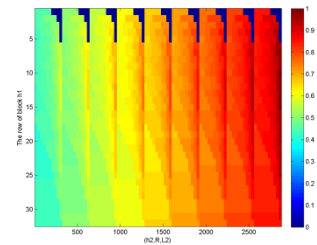
(*f*) $L_1$=6

(*g*) $L_1$=7

(*h*) $L_1$=8

(*i*) $L_1$=9

Fig. S16. The logarithm of BlockEnergy of ORL dataset in relation with $h_1$, $h_2$, $R$, $L_1$, $L_2$, where $h_1 \in \{1:1:32\}$, $h_2 = \lfloor nh_1/m \rfloor$, $R \in \{0:0.1:0.9\}$, $L_1 \in \{1:1:9\}$, $L_2 \in \{1:1:9\}$.

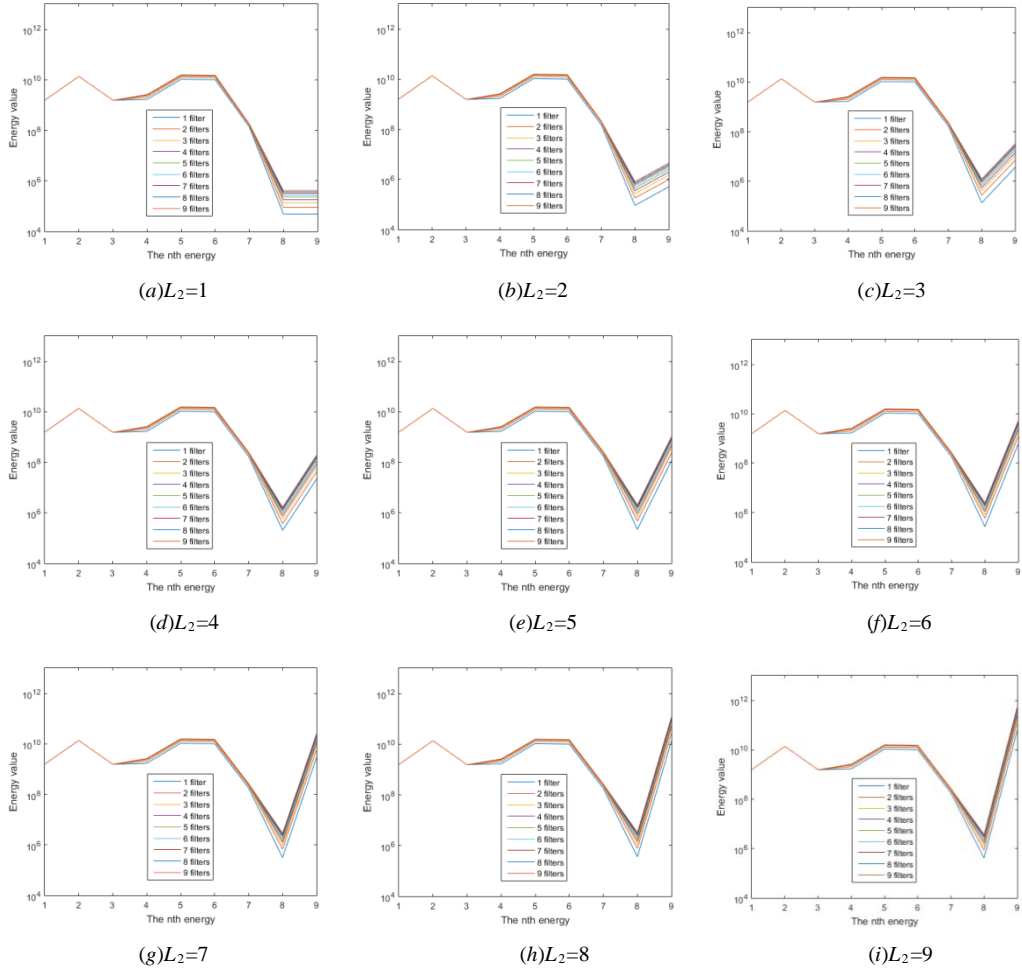

Fig. S17. The energy values of every stages of the first part of PCANet-2 versus the *n*th energy (ORL dataset). The parameters are $L_1$, $L_2 \in \{1:1:9\}$.

Table S9. The energy and energy ratio of BlockEnergy corresponding to $R \in \{0:0.1:0.9\}$. The parameters are $L_1 = L_2 = 6$, $h_1=h_2=8$. (ORL database)

| Overlap ratio $R$ | 0 | 0.1 | 0.2 | 0.3 | 0.4 | 0.5 | 0.6 | 0.7 | 0.8 | 0.9 |
|---|---|---|---|---|---|---|---|---|---|---|
| The $i$th filter | 1st | 2nd | 3rd | 4th | 5th | 6th | 7th | 8th | 9th | 10th |
| The energy of corresponding output image ($\times 10^{10}$) | 2.9905 | 0.8018 | 0.8018 | 0.3364 | 0.2028 | 0.1474 | 0.1031 | 0.1031 | 0.0676 | 0.0658 |
| Energy sum ($\times 10^{10}$) | 2.9905 | 3.7923 | 4.5941 | 4.9305 | 5.1333 | 5.2807 | 5.3838 | 5.4868 | 5.5545 | 5.6203 |
| The ratio of energy | 0.5321 | 0.6748 | 0.8174 | 0.8773 | 0.9133 | 0.9396 | 0.9579 | 0.9763 | 0.9883 | 1.0000 |

Table S10. The energy ratio and the eigenvalue ratio of 9 filters (the 1st and the 2nd PCAs). (ORL database).

| Parameters | The $n$th filter | Energy values ($\times 10^9$) | Energy Ratio | Eigenvalue ($\times 10^4$) | Eigenvalue sum ($\times 10^4$) | Eigenvalue Ratio | Error Rate |
|---|---|---|---|---|---|---|---|
| $h_1=h_2=8$ $R=0.5$ $L_2=1$ $L_1 \in \{1:1:9\}$ | 1 | 1.6843 | 0.6458 | 1.1939 | 1.1939 | 0.6289 | 0.3781 |
| | 2 | 2.0751 | 0.7956 | 0.3111 | 1.5050 | 0.7928 | 0.2219 |
| | 3 | 2.3443 | 0.8988 | 0.2139 | 1.7190 | 0.9055 | 0.1969 |
| | 4 | 2.4776 | 0.9500 | 0.0778 | 1.7967 | 0.9465 | 0.1625 |
| | 5 | 2.5587 | 0.9811 | 0.0503 | 1.8471 | 0.9730 | 0.1656 |
| | 6 | 2.5825 | 0.9902 | 0.0241 | 1.8712 | 0.9857 | 0.1594 |
| | 7 | 2.5929 | 0.9942 | 0.0110 | 1.8822 | 0.9915 | 0.1656 |
| | 8 | 2.6027 | 0.9979 | 0.0102 | 1.8925 | 0.9969 | 0.1594 |
| | 9 | 2.6081 | 1.0000 | 0.0059 | 1.8984 | 1.0000 | 0.1719 |
| $h_1=h_2=8$ $R=0.5$ $L_1=6$, $L_2 \in \{1:1:9\}$ | 1 | 0.17857 | 0.6716 | 2.0620 | 2.0620 | 0.6757 | 0.1594 |
| | 2 | 0.21460 | 0.8071 | 0.4401 | 2.5021 | 0.8200 | 0.1219 |
| | 3 | 0.23749 | 0.8932 | 0.2992 | 2.8013 | 0.9180 | 0.1031 |
| | 4 | 0.25336 | 0.9529 | 0.1237 | 2.9250 | 0.9586 | 0.0938 |
| | 5 | 0.26303 | 0.9893 | 0.0756 | 3.0007 | 0.9833 | 0.0781 |
| | 6 | 0.26487 | 0.9962 | 0.0304 | 3.0311 | 0.9933 | 0.0656 |
| | 7 | 0.26537 | 0.9981 | 0.0092 | 3.0403 | 0.9963 | 0.0469 |
| | 8 | 0.26578 | 0.9996 | 0.0089 | 3.0492 | 0.9993 | 0.0531 |
| | 9 | 0.26587 | 1.0000 | 0.0023 | 3.0515 | 1.0000 | 0.0656 |